\documentclass{article}
\PassOptionsToPackage{square, numbers, compress,sort}{natbib}
\usepackage[final]{neurips_2020}

\usepackage[utf8]{inputenc} 
\usepackage{amsbsy}  
\usepackage[T1]{fontenc}    
\usepackage{amsmath,amsfonts,amsthm,amssymb}       
\usepackage[hidelinks]{hyperref}       
\usepackage[capitalise]{cleveref}
\usepackage{color}
\usepackage{url}            
\usepackage{booktabs}       
\usepackage[algo2e,ruled]{algorithm2e}
\usepackage{nicefrac}       
\usepackage{microtype}      
\usepackage[pdftex]{graphicx}
\usepackage{subcaption}
\usepackage{mwe}
\usepackage{times}
\usepackage{titlesec} 
\usepackage{multirow}
\usepackage[inline]{enumitem}
\usepackage{hhline}  
\usepackage{aliascnt} 
\usepackage{todonotes} 
\usepackage{wrapfig}
\usepackage{mathtools} 
\usepackage{bm}

\newtheorem{observation}{Observation}
\newtheorem{property}{Property}
\newtheorem{theorem}{Theorem}
\newtheorem{corollary}{Corollary}
\newtheorem{lemma}{Lemma}

\newlength\inputlen

\crefname{equation}{Eq.}{Eqs.}
\crefname{figure}{Fig.}{Figs.}
\crefname{section}{Sec.}{Secs.}
\crefname{subsection}{Sec.}{Secs.}
\crefname{theorem}{Thm.}{Thms.}
\crefname{appendix}{Appx.}{Appx.}
\crefname{lemma}{Lemma}{Lemmas}
\crefname{algocf}{Alg.}{Algs.}
\Crefname{algocf}{Algorithm}{Algorithms}

\newif\ifcomments
\commentstrue
\ifcomments\newcommand{\comments}[1]{#1}\else\newcommand{\comments}[1]{}\fi
\newcommand{\new}[1]{{#1}}

\definecolor{clrgp}{rgb}{.9,0,.9}
\definecolor{gray}{rgb}{0.41, 0.41, 0.41}
\definecolor{forestgreen}{rgb}{0.13, 0.55, 0.13}

\titleformat*{\subparagraph}{\normalfont\itshape}
\renewcommand{\paragraph}[1]{\vspace{0.20ex}\noindent\textbf{#1}}
\renewcommand{\subparagraph}[1]{\vspace{0.20ex}\noindent\textit{#1}}

\usepackage{xcolor}
\definecolor{dark-red}{rgb}{0.4,0.15,0.15}
\definecolor{dark-blue}{rgb}{0.15,0.15,0.4}
\definecolor{medium-blue}{rgb}{0,0,0.5}
\hypersetup{
   colorlinks, linkcolor={dark-blue},
   citecolor={dark-blue}, urlcolor={medium-blue}
}

\setlist[enumerate,1]{label={\arabic*)}}

\newcommand{\titl}{Fast Matrix Square Roots with Applications to Gaussian Processes and Bayesian Optimization}

\newcommand{\authorinfo}{
  Geoff Pleiss \\
  Columbia University \\
  gmp2162@columbia.edu
  \And
  Martin Jankowiak \\
  The Broad Institute \\
  mjankowi@broadinstitute.org
  \And
  David Eriksson\thanks{This work was conducted while David Eriksson was at Uber AI.}\\
  Facebook \\
  deriksson@fb.com \\
  \And
  Anil Damle \\
  Cornell University \\
  damle@cornell.edu
  \And
  Jacob R. Gardner \\
  University of Pennsylvania \\
  jacobrg@seas.upenn.edu
}

\title{\titl}
\author{\authorinfo}

\DeclareMathOperator*{\argmin}{arg\,min}
\DeclareMathOperator*{\expectedvalue}{\mathbb{E}}

\DeclareMathOperator*{\trace}{Tr}
\DeclareMathOperator*{\kltext}{KL}

\newcommand{\bigo}[1]{\ensuremath{\mathchoice{\mathcal O \! \left( #1 \right)}}{\mathcal O(#1)}{\mathcal O(#1)}{\mathcal O(#1)}}

\newcommand{\reals}{\ensuremath{\mathbb{R}}}

\newcommand{\normaldist}[2]{\ensuremath{\mathchoice{\mathcal{N} \left( #1, #2 \right) }{ \mathcal{N}(#1,#2) }{ \mathcal{N}(#1,#2) }{}}}

\newcommand{\Evover}[2]{\ensuremath{\expectedvalue_{#1} \left[ #2 \right]}}

\newcommand{\kl}[2]{\ensuremath{\kltext \left[ \: #1 \Vert #2 \: \right]}}

\newcommand{\NN}{\mathcal{N}}

\newcommand{\tr}[1]{\ensuremath{\mathchoice{\trace \left( #1 \right)}{\trace ( #1 )}{}{}}}

\newcommand{\intd}[1]{\,\mathrm{d}{#1}}

\newif\ifboldmatrix
\boldmatrixtrue
\newcommand{\mathbcal}[1]{\pmb{\mathcal{#1}}}
\ifboldmatrix\newcommand{\boldmatrix}[1]{\mathbf{#1}}\else\newcommand{\boldmatrix}[1]{#1}\fi
\ifboldmatrix\newcommand{\boldgreekmatrix}[1]{\boldsymbol{#1}}\else\newcommand{\boldgreekmatrix}[1]{#1}\fi
\ifboldmatrix\newcommand{\boldscriptmatrix}[1]{\mathbcal{#1}}\else\newcommand{\boldscriptmatrix}[1]{#1}\fi
\newcommand{\ba}{\ensuremath{\mathbf{a}}}

\newcommand{\bb}{\ensuremath{\mathbf{b}}}

\newcommand{\bc}{\ensuremath{\mathbf{c}}}
\newcommand{\bd}{\ensuremath{\mathbf{d}}}
\newcommand{\be}{\ensuremath{\mathbf{e}}}
\newcommand{\boeta}{\ensuremath{\boldsymbol{\eta}}}
\newcommand{\bepsilon}{\ensuremath{\boldsymbol{\epsilon}}}

\newcommand{\bk}{\ensuremath{\mathbf{k}}}

\newcommand{\bmm}{\ensuremath{\mathbf{m}}}
\newcommand{\bmu}{\ensuremath{\boldsymbol{\mu}}}

\newcommand{\bq}{\ensuremath{\mathbf{q}}}
\newcommand{\bqu}{\ensuremath{\mathbcal{Q}}}
\newcommand{\br}{\ensuremath{\mathbf{r}}}

\newcommand{\bt}{\ensuremath{\mathbf{t}}}
\newcommand{\btheta}{\ensuremath{\boldsymbol{\theta}}}
\newcommand{\bu}{\ensuremath{\mathbf{u}}}
\newcommand{\bv}{\ensuremath{\mathbf{v}}}

\newcommand{\bx}{\ensuremath{\mathbf{x}}}
\newcommand{\by}{\ensuremath{\mathbf{y}}}
\newcommand{\bz}{\ensuremath{\mathbf{z}}}
\newcommand{\bzero}{\ensuremath{\mathbf{0}}}
\newcommand{\bone}{\ensuremath{\mathbf{1}}}
\newcommand{\bA}{\ensuremath{\boldmatrix{A}}}

\newcommand{\bEta}{\ensuremath{\boldmatrix{H}}}

\newcommand{\bFS}{\ensuremath{\boldscriptmatrix{F}}}
\newcommand{\bI}{\ensuremath{\boldmatrix{I}}}
\newcommand{\bK}{\ensuremath{\boldmatrix{K}}}
\newcommand{\bL}{\ensuremath{\boldmatrix{L}}}

\newcommand{\bP}{\ensuremath{\boldmatrix{P}}}

\newcommand{\bQ}{\ensuremath{\boldmatrix{Q}}}
\newcommand{\bQU}{\ensuremath{\boldscriptmatrix{Q}}}
\newcommand{\bR}{\ensuremath{\boldmatrix{R}}}
\newcommand{\bS}{\ensuremath{\boldmatrix{S}}}

\newcommand{\bTheta}{\ensuremath{\boldgreekmatrix{\Theta}}}
\newcommand{\bT}{\ensuremath{\boldmatrix{T}}}

\newcommand{\bX}{\ensuremath{\boldmatrix{X}}}

\newcommand{\bZ}{\ensuremath{\boldmatrix{Z}}}

\newcommand{\bxtest}{\ensuremath{\bx^*}}

\newcommand{\bXtest}{\ensuremath{\bX^*}}

\newcommand{\bmeantest}{\ensuremath{\bmu^*}}

\newcommand{\Covtest}{\ensuremath{\textbf{COV}^*}}
\newcommand{\ameantest}[1]{\ensuremath{\mu^*_\text{aprx} \left( #1 \right)}}
\newcommand{\avartest}[1]{\ensuremath{\text{Var}^*_\text{aprx} \left( #1 \right)}}

\newcommand{\loglik}{\ensuremath{\mathcal L}}

\newcommand{\mvm}[1]{\ensuremath{\xi ( #1 )}}

\newcommand{\bLam}{\bm{\Lambda}}

\begin{document}
\maketitle

\begin{abstract}
  Matrix square roots and their inverses arise frequently in machine learning, e.g., when sampling from high-dimensional Gaussians $\normaldist{\bzero}{\bK}$ or ``whitening'' a vector $\bb$ against covariance matrix $\bK$.
  While existing methods typically require $\bigo{N^3}$ computation, we introduce a highly-efficient quadratic-time algorithm for computing $\bK^{1/2} \bb$, $\bK^{-1/2} \bb$, and their derivatives through \emph{matrix-vector multiplication} (MVMs).
  Our method combines Krylov subspace methods with a rational approximation and typically achieves $4$ decimal places of accuracy with fewer than $100$ MVMs.
  Moreover, the backward pass requires little additional computation.
  We demonstrate our method's applicability on matrices as large as $50,\!000 \times 50,\!000$---well beyond traditional methods---with little approximation error.
  Applying this increased scalability to variational Gaussian processes, Bayesian optimization, and Gibbs sampling results in more powerful models with higher accuracy.
  \new{
  In particular, we perform variational GP inference with up to $10,\!000$ inducing points and perform Gibbs sampling on a $25,\!000$-dimensional problem.
  }
\end{abstract}

\section{Introduction}

High-dimensional Gaussian distributions arise frequently in machine learning, especially in the context of Bayesian modeling.
For example, the prior of Gaussian process models is given by a multivariate Gaussian distribution
$\normaldist{\bzero}{\bK}$ governed by an $N \times N$ symmetric positive definite kernel matrix $\bK$.
Historically, $\bigo{N^3}$ computation and $\bigo{N^2}$ memory requirements have limited the tractability of inference for high-dimensional Gaussian latent variable models.

A growing line of research aims to reformulate many common covariance matrix operations---such as linear solves and log determinants---as iterative optimizations involving \emph{matrix-vector multiplications} (MVMs) \citep[e.g.][]{aune2013iterative,cutajar2016preconditioning,charlier2020kernel,gardner2018gpytorch,wang2019exact}.
MVM approaches have two primary advantages:
1) the covariance matrix need not be explicitly instantiated (so only $\bigo{N}$ memory is required) \cite{cutajar2016preconditioning,wang2019exact,charlier2020kernel}; and
  2) MVMs utilize GPU acceleration better than direct methods like Cholesky \cite{aune2013iterative,gardner2018gpytorch}.
Thus MVM methods can be scaled to much larger covariance matrices.

In this paper, we propose an MVM method that addresses a common computational bottleneck for high-dimensional Gaussians: computing $\bK^{\pm 1/2} \bb$.
This operation occurs frequently in Gaussian process models and inverse problems.
For example, if $\bb \sim \normaldist{\bzero}{\bI}$, then $\bK^{\frac 1 2} \bb \sim \normaldist{\bzero}{\bK}$.
This operation appears frequently in Bayesian optimization \citep[e.g.][]{thompson1933likelihood,frazier2009knowledge,hernandez2014predictive,wang2017max} and Gibbs sampling \citep[e.g.][]{besag1991bayesian,bardsley2012mcmc,griffin2017hierarchical}.
$\bK^{-\frac 1 2} \bb$ can be used to project parameters into a ``whitened'' coordinate space \cite{kuss2005assessing,matthews2017scalable}---a transformation that accelerates the convergence of variational Gaussian process approximations.
To make these computations more efficient and scalable, we make the following contributions:
\begin{itemize}
  \item We introduce a MVM approach for computing $\bK^{\pm 1 / 2} \bb$.
    The approach uses an insight from \citet{hale2008computing} that expresses the matrix square root as a sum of shifted matrix inverses.

  \item To efficiently compute these shifted inverses, we leverage a modified version of the MINRES algorithm \cite{paige1975solution} that performs \emph{multiple shifted solves} through a single iteration of MVMs.
    We demonstrate that, surprisingly, {\bf multi-shift MINRES (msMINRES)} convergence can be accelerated with a \emph{single} preconditioner despite the presence of multiple shifts.
    Moreover, msMINRES only requires $\bigo{N}$ storage when used in conjunction with partitioned MVMs \cite{wang2019exact,charlier2020kernel}.
    Achieving 4 or 5 decimal places of accuracy typically requires \emph{fewer than 100 matrix-vector multiplications}, which can be highly accelerated through GPUs.

  \item We derive a scalable backward pass for $\bK^{\pm 1/2} \bb$  that enables our approach to be used as part of learning and optimization.

  \item We apply our $\bK^{-1/2} \bb$ and $\bK^{1/2} \bb$ routines to three applications:
      1) variational Gaussian processes with up to $M = 10^4$ inducing points (where we additionally introduce a $\bigo{M^2}$ MVM-based natural gradient update);
      2) sampling from Gaussian process posteriors in the context of Bayesian optimization with up to $50,\!000$ candidate points;
      and 3) an image reconstruction task where we perform Gibbs sampling in $25,\!600$ dimensions.
\end{itemize}

\new{
Code examples for the GPyTorch framework are available at \url{bit.ly/ciq_svgp} and \url{bit.ly/ciq_sampling}.
}

\section{Background}

\paragraph{Existing Methods for Sampling and Whitening}
typically rely on the Cholesky factorization: $\bK = \bL \bL^\top$, where $\bL$ is lower triangular.
Though $\bL$ is not a square root of $\bK$, $\bL\bb$ is equivalent to $\bK^{1/2} \bb$ up to an orthonormal rotation.
Therefore, $\bL\bepsilon$, $\bepsilon \sim \normaldist{\bzero}{\bI}$ can be used to draw samples from from $\normaldist{\bzero}{\bK}$ and $\bL^{-1} \bb$ can be used to ``whiten'' the vector $\bb$.
However, the Cholesky factor requires $\bigo{N^3}$ computation and $\bigo{N^2}$ memory for an $N \times N$ covariance matrix $\bK$.
To avoid this large complexity, randomized algorithms \cite{rahimi2008random,mutny2018efficient}, low-rank/sparse approximations \cite{hensman2017variational,pleiss2018constant,wilson2020efficiently}, or alternative distributions \citep{wang2017max} are often used to approximate the sampling and whitening operations.

\paragraph{Krylov Subspace Methods}
are a family of iterative algorithms for computing functions of matrices applied to vectors $f(\bK) \bb$ \citep[e.g.][]{schneider2001krylov,saad2003iterative,van2003iterative}.
Crucially, $\bK$ is only accessed through \emph{matrix-vector multiplication} (MVM), which is beneficial for extremely large matrices that cannot be explicitly computed in memory.
All Krylov algorithms share the same basic structure: each iteration $j$ produces an estimate $\bc_j \approx f(\bK) \bb$ which falls within the $j^\text{th}$ \emph{Krylov subspace} of $\bK$ and $\bb$:
\begin{equation}
  \bc_j \in \mathcal{K}_j (\bK, \bb) = \text{span} \left\{ \bb, \:\: \bK \bb, \:\: \bK^2 \bb, \:\: \ldots, \:\: \bK^{j-1} \bb \right\}.
  \label{eqn:krylov}
\end{equation}
Each iteration expands the Krylov subspace by one vector, requiring a single matrix-vector multiplication with $\bK$.
Many Krylov methods, such as linear conjugate gradients, can be reduced to computationally efficient vector recurrences.
Krylov methods are exact after $N$ iterations, though most methods offer extremely accurate solutions in $J \ll N$ iterations.
There has been growing interest in applying Krylov methods to large-scale kernel methods \citep{aune2013iterative,aune2014parameter,charlier2020kernel,chow2014preconditioned,cunningham2008fast,cutajar2016preconditioning,gardner2018product,gardner2018gpytorch,murray2009gaussian,pleiss2018constant,saad2003iterative,simpson2008fast,simpson2013scalable,wilson2015kernel}, especially due to their memory efficiency and amenability to GPU acceleration.

\section{Contour Integral Quadrature (CIQ) via Matrix-Vector Multiplication}
\label{sec:ciq_method}

In this section we develop an MVM method to compute $\bK^{-1/2} \bb$ and $\bK^{1/2} \bb$ for sampling and whitening.
Our approach scales better than existing methods (e.g.~Cholesky) by:
1) reducing computation from $\bigo{N^3}$ to $\bigo{N^2}$;
2) reducing memory from $\bigo{N^2}$ to $\bigo{N}$;
3) more effectively using GPU acceleration; and
4) affording an efficient gradient computation.

\paragraph{Contour Integral Quadrature (CIQ).}
A standard result from complex analysis is that $\bK^{-1/2}$ can be expressed through Cauchy's integral formula:
$
	\bK^{-1 / 2} = \frac{1}{2 \pi i} \oint_\Gamma \tau^{- 1 / 2} \left( \tau \bI - \bK \right)^{-1} \intd \tau,
$
where $\Gamma$ is a closed contour in the complex plane that winds once around the spectrum of $\bK$ \citep{davies2005computing,hale2008computing,higham2008functions}.
Applying a numerical quadrature scheme to the contour integral yields the rational approximations
\begin{equation}
	\bK^{-\frac 1 2} \approx \sum_{q=1}^Q w_q \left( t_q \bI + \bK \right)^{-1}
  \quad \text{and} \quad
	\bK^{\frac 1 2} \approx \bK \sum_{q=1}^Q w_q \left( t_q \bI + \bK \right)^{-1},
	\label{eqn:contour_integral_quad}
\end{equation}
where the weights $w_q$ encapsulate the normalizing constant, quadrature weights, and the $t_q^{-\frac 1 2}$ terms.
\citet{hale2008computing} introduce a real-valued quadrature strategy based on a change-of-variables formulation (described in \cref{app:quadrature})
that converges extremely rapidly---often achieving full machine precision with only $Q \approx 20$ quadrature points.
For the remainder of this paper, applying~\cref{eqn:contour_integral_quad} to compute $\bK^{\pm1/2} \bb$  will be referred to as {\bf Contour Integral Quadrature (CIQ)}.

\subsection{An Efficient Matrix-Vector Multiplication Approach to CIQ with msMINRES.}
Using the quadrature method of \cref{eqn:contour_integral_quad} for whitening and sampling requires solving several shifted linear systems.
To compute the shifted solves required by~\cref{eqn:contour_integral_quad} we leverage a variant of the
minimum residuals algorithm (MINRES) developed by \citet{paige1975solution}. At step $j$
MINRES approximates $\bK^{-1} \bb$ by the vector within the Krylov subspace $\bc \in \mathcal{K}_j(\bK, \bb)$ that minimizes the residual $\Vert \bK \bc - \bb \Vert_2$.

\paragraph{msMINRES for multiple shifted solves.}
To efficiently compute all the shifted solves, we leverage techniques~\citep[e.g.][]{datta1991arnoldi,freund1990conjugate,frommer1998restarted,cundy2009numerical,meerbergen2003solution} that exploit the shift-invariance property of Krylov subspaces: i.e.~$\mathcal{K}_J(\bK, \bb) = \mathcal{K}_J(t \bI +  \bK, \bb)$.
We introduce a variant to MINRES, which we refer to as {\bf multi-shift MINRES} or {\bf msMINRES}, that re-uses the same Krylov subspace vectors $[\bb, \: \bK \bb, \: \ldots, \: \bK^{J-1} \bb]$ for all shifted solves $(t \bI + \bK)^{-1} \bb$.
In other words, using msMINRES we can get all $(t_q \bI + \bK)^{-1} \bb$ \emph{essentially for free}, i.e. only requiring $J$ MVMs for the Krylov subspace $\mathcal{K}_J(\bK, \bb)$.
As with standard MINRES, the msMINRES procedure for computing $(t_q \bI + \bK)^{-1}$ from $[\bb, \: \bK \bb, \: \ldots, \: \bK^{J-1} \bb]$ can be reduced to a simple vector recurrence (see \cref{app:minres} for details).

\subsection{Computational Complexity and Convergence Analysis of msMINRES-CIQ}
Pairing \cref{eqn:contour_integral_quad} with msMINRES is an efficient algorithm for computing $\bK^{1/2} \bb$ and $\bK^{-1/2} \bb$.
\cref{alg:ciq} (see Appendix) summarizes this approach; below we highlight its computational properties:

\label{sec:convergence}

\begin{property}[Computation/Memory of msMINRES-CIQ]
  $J$ iterations of msMINRES requires exactly $J$ MVMs with the input matrix $\bK$,
  regardless of the number of quadrature points $Q$.
  The resulting runtime of msMINRES-CIQ is $\bigo{ J \mvm{\bK}}$, where $\mvm{\bK}$ is the time to perform an MVM with $\bK$.
  The memory requirement is $\bigo{ QN }$ in addition to what is required to store $\bK$.
  \label{prop:msminres}
\end{property}
For arbitrary positive semi-definite $N \! \times \! N$ matrices, the runtime of msMINRES-CIQ is $\bigo{J N^2}$, where often $J \ll N$.
Performing the MVMs in a map-reduce fashion \cite{wang2019exact,charlier2020kernel} avoids explicitly forming $\bK$, which results in $\bigo{QN}$ total memory.
\new{
This is in contrast to Cholesky, which produces an artifact that requires $\bigo{N^2}$ memory.
}
Below we bound the error of msMINRES-CIQ:
\begin{theorem}
  Let $\bK \succ 0$ and $\bb$ be inputs to msMINRES-CIQ, producing $\ba_J \approx \bK^{1/2} \bb$ after $J$ iterations with $Q$ quadrature points.
  The difference between $\ba_J$ and $\bK^{1/2} \bb$ is bounded by:
  \begin{equation*}
    \left\Vert \ba_J - \bK^{\frac 1 2} \bb \right\Vert_2
    \leq
    \overbracket{
      \bigo{\exp\left( -\tfrac  {2 Q \pi^2}{\log \kappa(\bK) + 3} \right)}
    }^{\text{Quadrature error}}
    +
    \overbracket{
      \tfrac{ 2 Q \log \left( 5 \sqrt{\kappa(\bK)} \right) \kappa(\bK) \sqrt{\lambda_\text{min}} }{\pi}
      \left( \tfrac{ \sqrt{\kappa(\bK)} - 1}{ \sqrt{\kappa(\bK)} + 1} \right)^{J-1}
      \left\Vert \bb \right\Vert_2.
    }^{\text{msMINRES error}}
  \end{equation*}
  where $\lambda_\text{max},\lambda_{\text{min}}$ are the max and min eigenvalues of $\bK$, and $\kappa(\bK) \equiv \tfrac{\lambda_\text{max}}{\lambda_\text{min}}$ is the condition number.
  \label{thm:ciq_convergence}
\end{theorem}
For $\ba'_J \approx \bK^{-1/2} \bb$, the bound incurs an additional factor of $1/\lambda_\text{min}$.
(See \cref{app:proofs} for proofs.)
\cref{thm:ciq_convergence} suggests that error in computing $(t_q \bI + \bK)^{-1}\bb$  will be the primary source of error as the quadrature error decays rapidly with $Q$. In many of our applications the rapid convergence of Krylov subspace methods for linear solves is well established, allowing for accurate solutions if desired.
For covariance matrices up to $N=50,\!000$, often $Q=8$ and $J\leq100$ suffices for 4 decimal places of accuracy and $J$ can be further reduced with preconditioning (see \cref{sec:empirical,app:preconditioner}).

\subsection{Efficient Vector-Jacobi Products for Backpropagation}

In certain applications, such as variational Gaussian process inference, we have to compute gradients of the $\bK^{-1/2} \bb$ operation.
This requires the vector-Jacobian product $\bv^\top ( \partial \bK^{-1/2} \bb / \partial \bK)$, where $\bv$ is the back-propagated gradient.
The form of the Jacobian is the solution to a Lyapunov equation, which requires expensive iterative methods or solving a $N^2 \times N^2$ Kronecker sum $(\bK^{1/2} \oplus \bK^{1/2})^{-1}$.
Both of these options are much slower than the forward pass and are impractical for large $N$.
Fortunately, our quadrature formulation affords a computationally efficient approximation to this vector-Jacobian product.
If we back-propagate directly through each term in \cref{eqn:contour_integral_quad}, we have
\begin{align}
  \bv^\top \left( \frac{ \partial \bK^{-1/2} \bb }{ \partial \bK } \right)
\approx
 - \frac{1}{2} \sum_{q=1}^{Q} w_q \left( t_q \bI + \bK \right)^{-1} \left( \bv \bb^\top + \bb \bv^\top \right) \left( t_q \bI + \bK \right)^{-1}.
  \label{eqn:ciq_deriv}
\end{align}
Since the forward pass computes the solves with $\bb$, the only additional work needed for the backward pass is computing the shifted solves $(t_q \bI + \bK)^{-1} \bv$, which can be computed with another call to the msMINRES algorithm.
Thus the backward pass takes only $\bigo{J \mvm{\bK}}$ (e.g.~$\bigo{J N^2}$) time.

\new{
\subsection{Preconditioning}
Preconditioners are commonly applied to Krylov subspace methods like MINRES to improve the condition number $\kappa(\bK)$ and accelerate convergence.
However, standard preconditioning techniques do not apply to msMINRES, as each shifted system $\bK + t_q \bI$ requires its own preconditioner (see \cref{app:preconditioner} for details).
Each separately preconditioned system would require separate MVMs, defeating the efficiency of msMINRES.
Nevertheless, we can use a single preconditioner to compute rotationally-equivalent solutions to $\bK^{\pm1/2} \bb$.
If $\bP \approx \bK$ is a preconditioner matrix, we note that:
\[
	\bK \bP^{-\frac 1 2} (\bP^{-\frac 1 2} \bK \bP^{-\frac 1 2})^{-\frac 1 2} \bb, \qquad
	\bP^{-\frac 1 2} (\bP^{-\frac 1 2} \bK \bP^{-\frac 1 2})^{-\frac 1 2} \bb
\]
are equivalent to $\bK^{1/2} \bb$ and $\bK^{-1/2} \bb$ (respectively) up to an orthonormal rotation (see \cref{app:preconditioner}).
We can use msMINRES-CIQ to compute the $(\bP^{-1/2} \bK \bP^{-1/2})^{-1/2} \bb$ terms.
Crucially, the convergence now depends on the condition number of $\bP^{-1/2} \bK \bP^{-1/2}$, rather than that of $\bK$.
}

\subsection{Related Work}
Other Krylov methods for $\bK^{1/2} \bb$ and $\bK^{-1/2} \bb,$ often via polynomial approximations~\citep[e.g.][]{higham2008functions}, have been explored.
\citet{chow2014preconditioned} compute $\bK^{1/2} \bb$ via a preconditioned Lanczos algorithm.
Unlike msMINRES, however, they require storage of the entire Krylov subspace. Moreover this approach does not afford a simple gradient computation.
\new{
\citeauthor{frommer2014efficient} \cite{frommer2014efficient,frommer2014convergence} apply a similar Krylov/quadrature approach to a broad class of matrix functions.
}
More similar to our work is \cite{aune2013iterative,aune2014parameter}, which uses the quadrature formulation of \cref{eqn:contour_integral_quad} in conjunction with a shifted conjugate gradients solver.
We expand upon their method by:
  1) introducing a simple gradient computation;
  2) proving a convergence guarantee; and
  3) enabling the use of simple preconditioners (see \cref{app:preconditioner}).

\section{Benchmarking msMINRES-CIQ}
\label{sec:empirical}

\begin{figure}[t!]
	\centering
	\includegraphics[width=\textwidth]{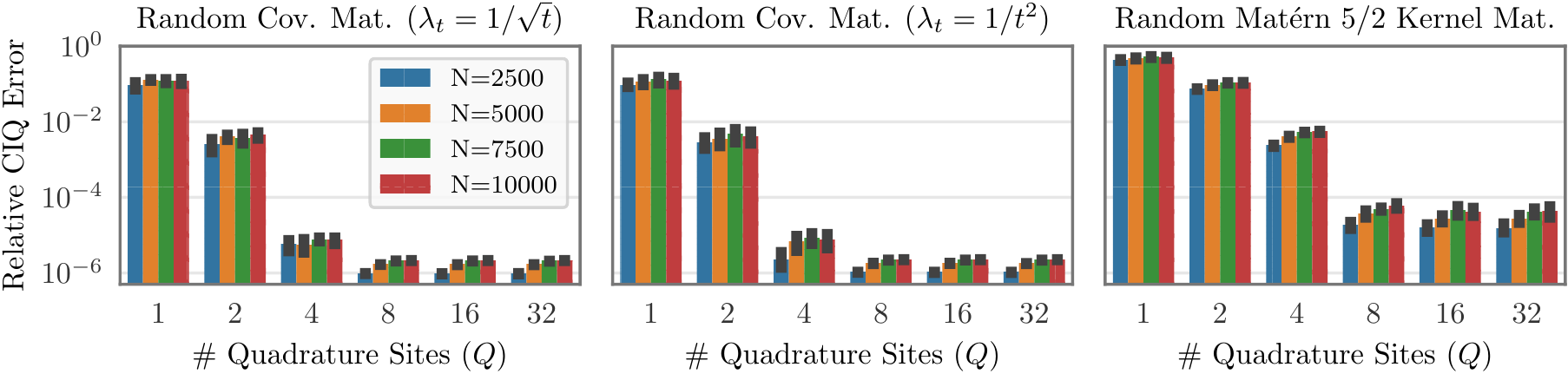}
  \caption{
    msMINRES-CIQ relative error when computing $\bK^{1/2} \bb$ as a function of number of quadrature sites $Q$.
    We test random matrices with eigenvalues that scale as $\lambda_t = 1/\sqrt{t}$ (left) and $\lambda_t = 1/{t}^2$ (middle),
    as well as Mat\'ern kernels (right).
    In all cases $Q\!=\!8$ achieves $<10^{-4}$ error.
    \new{
    The error levels out at roughly $10^{-4}$ or $10^{-5}$, which corresponds to the msMINRES tolerance.
    }
		msMINRES is stopped after achieving a relative residual of $10^{-4}$ or $J=400$ iterations.
  }
  \label{fig:quad_error}
\end{figure}

\begin{figure}[t!]
	\centering
	\includegraphics[width=0.36\textwidth]{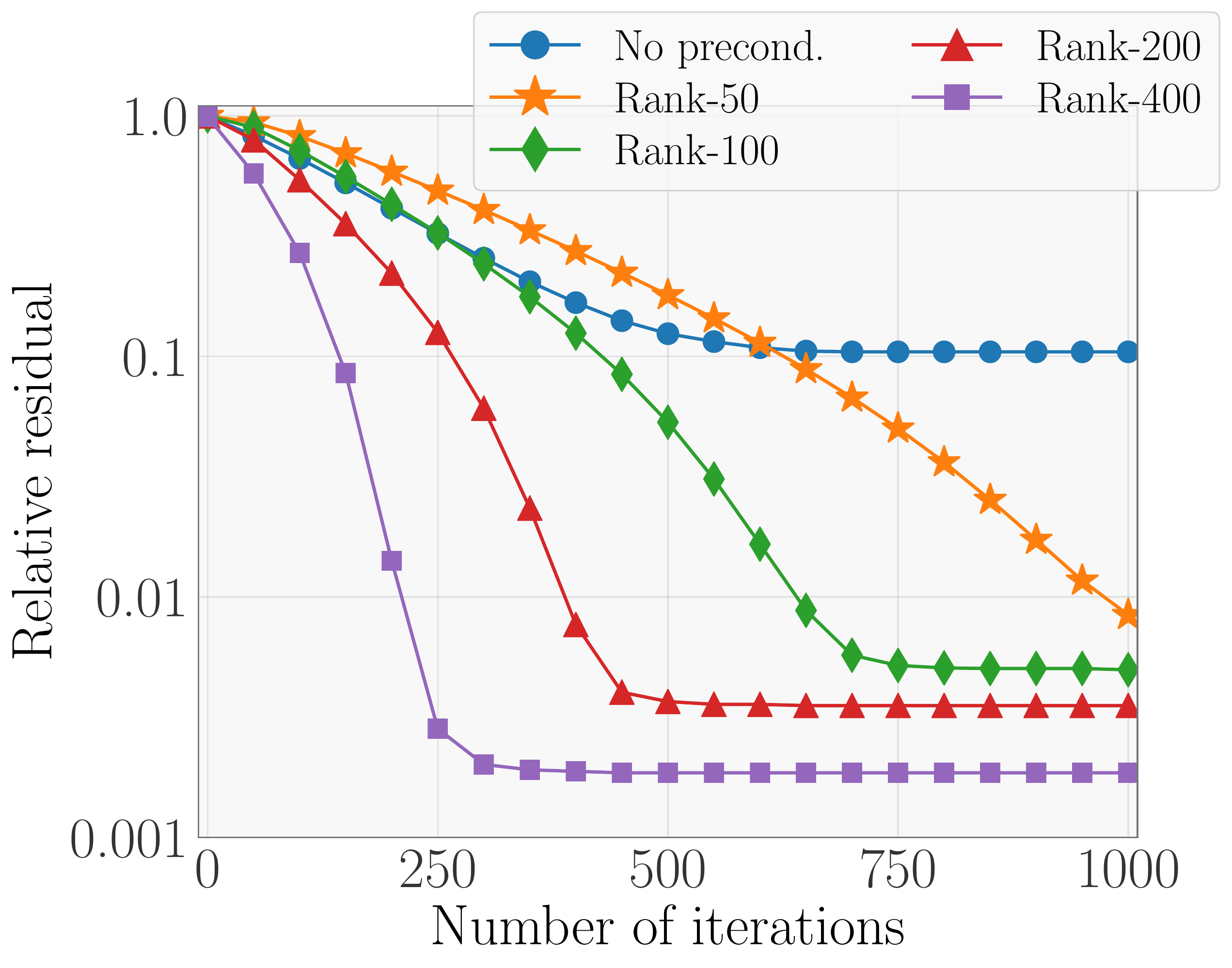}
  \quad
	\includegraphics[width=0.6\textwidth]{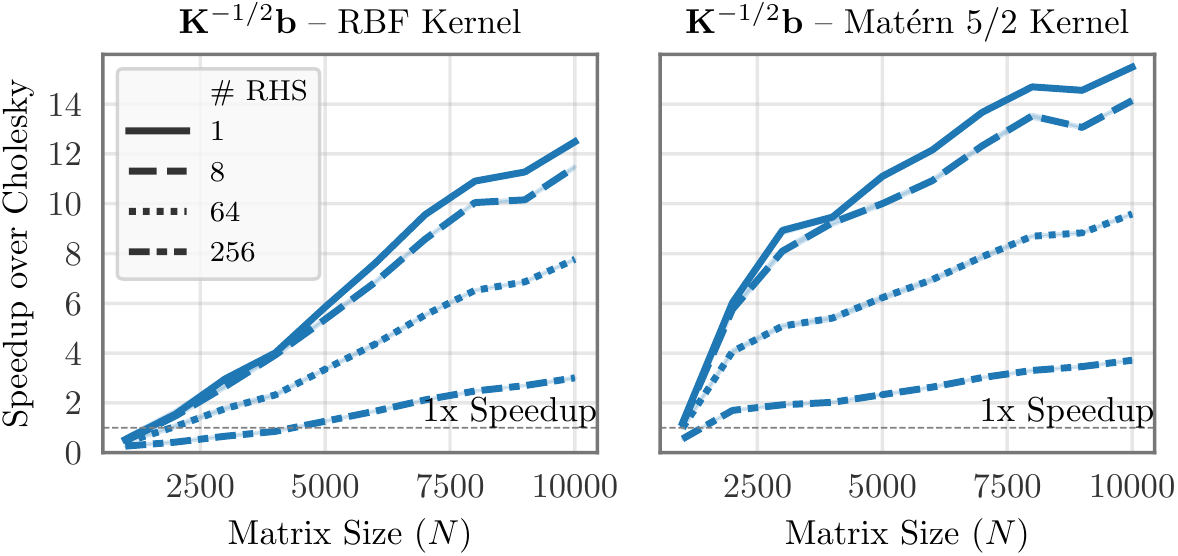}
  \caption{
    ({\bf Left}:) Effect of preconditioning on msMINRES-CIQ convergence while performing Bayesian optimization.
		Samples are drawn from the $N=50,\!000$ posterior covariance matrix of the ill-conditioned 6-dimensional Hartmann function (see \cref{sec:bayesopt_results}), using the pivoted Cholesky preconditioner \citep{gardner2018gpytorch}.
    ({\bf Middle/Right}:) Speedup of msMINRES-CIQ over Cholesky when computing forward/backward passes of $\bK^{-1/2} \bb$ with varying number of right-hand-sides $\bb$ (RHS).
  }
  \label{fig:precond_result}
\end{figure}

In this section we empirically measure the convergence and speedup of msMINRES-CIQ applied to several types of covariance matrices.

\paragraph{Convergence of msMINRES-CIQ.}
In \cref{fig:quad_error} we measure the relative error of computing $\bK^{1/2} \bb$ with msMINRES-CIQ on random matrices.\footnote{
	msMINRES is stopped after achieving a relative residual of $10^{-4}$ or after reaching $J=400$ iterations.
}
We vary
\begin{enumerate*}
  \item the number of quadrature points $Q$;
  \item the size of the matrix $N$; and
  \item the conditioning of the matrix.
\end{enumerate*}
The left and middle plots display results for matrices with spectra that decay as $\lambda_t = 1 / \sqrt{t}$ and $\lambda_t = 1 / t^2$, respectively.
The right plot displays results for one-dimensional Mat\'ern kernel matrices (formed with random data), which have near-exponentially decaying spectra.
Consequently, the $1 / \sqrt{t}$ matrices are relatively well-conditioned, while the Mat\'ern kernels are relatively ill-conditioned.
Nevertheless, in all cases CIQ achieves $10^{-4}$ relative error with only $Q=8$ quadrature points, regardless of the size of the matrix.
Additionally, \cref{app:additional_results} demonstrates that msMINRES-CIQ achieves orders of magnitude smaller error than approximation algorithms like randomized SVD \citep{halko2009finding} or random Fourier features \cite{rahimi2008random}.

To demonstrate the effect of preconditioning, we construct a posterior covariance matrix of size $N=50,\!000$ points on the $6$ dimensional Hartmann function (see \cref{sec:bayesopt_results} for a description).
We note that this problem is particularly ill-conditioned ($\kappa(\bK) \approx 10^8$), and thus represents an extreme test case.
\cref{fig:precond_result} (left) plots the convergence of msMINRES-CIQ (computing $\bK^{1/2} \bb$).
Without preconditioning, it is difficult to achieve relative residuals less than $0.1$.
Using the pivoted Cholesky preconditioner of \citet{gardner2018gpytorch}---a low-rank approximation of $\bK$---not only accelerates the convergence but also reduces the final residual.
With rank-200/rank-400 preconditioners, the final residual is cut by orders of magnitude, and msMINRES-CIQ converges $2\times$/$4\times$ faster.

\paragraph{Speedup over Cholesky.}
We compare the wall-clock speedup of msMINRES-CIQ over Cholesky in \cref{fig:precond_result} (middle/right) on RBF/Mat\'ern kernels.\footnote{
  $Q=8$.
  msMINRES is stopped after a residual of $10^{-4}$.
  Kernels are formed using data from the Kin40k dataset \citep{asuncion2007uci}.
  Timings are performed on a NVIDIA 1070 GPU.
}
We compute $\bK^{-1/2} \bb$ and its derivative on multiple right-hand-side (RHS) vectors.
As $N$ increases, msMINRES-CIQ incurs a larger speedup (up to $15\times$ faster than Cholesky).
This speedup is less pronounced when computing many RHSs simultaneously, as the cubic complexity of Cholesky is amortized across each RHS.
Nevertheless, msMINRES-CIQ is advantageous for matrices larger than $N=3,\!000$ even when simultaneously whitening $256$ vectors.

\section{Applications}

In previous sections we showed, both theoretically and empirically, that msMINRES-CIQ accurately computes $\bK^{\pm1/2} \bb$ while scaling better than traditional (Cholesky-based) methods.
In this section we demonstrate applications of this increased speed and scalability.
In particular, we show that using msMINRES-CIQ in conjunction with variational Gaussian processes, Bayesian optimization, and Gibbs sampling facilitates higher-fidelity models that can be applied to large-scale problems.
\subsection{Whitened Stochastic Variational Gaussian Processes}
\label{sec:variational_results}

As a first application, we demonstrate that the msMINRES-CIQ whitening procedure $\bK^{-1/2} \bb$ can increase the fidelity of {\bf stochastic variational Gaussian processes (SVGP)} \cite{hensman2013gaussian,hensman2015scalable,matthews2017scalable}.
These models are used for non-conjugate likelihoods (e.g.~binary classification) or for large datasets that do not fit into memory.
SVGP forms an approximate posterior
$
  p(f(\bx) \mid \bX, \by) \approx q(f(\bx)) = \Evover{q(\bu)}{ p\left( f(\bx) \mid \bu \right)},
$
where $\bu \in \reals^M$ are {inducing function values} (see \citep{hensman2015scalable,matthews2017scalable} for a detailed derivation).
$q \left( \bu \right)$ is a Gaussian variational distribution parameterized by mean $\bmm \in \reals^M$ and covariance $\bS \in \reals^{M \times M}$.
$\bmm$ and $\bS$ (as well as the model's kernel/likelihood hyperparameters) are chosen to maximize the variational ELBO:
\begin{align*}
  \loglik_\text{ELBO}\bigl\{ q(\bu) = \normaldist{\bmm}{\bS} \bigr\} &= \textstyle\sum_{i=1}^N \Evover{q(f(\bx^{(i)}))}{  \: \log p( y^{(i)} \mid f(\bx^{(i)}) ) \: } - \kl{ q(\bu) }{ p(\bu) }.
\end{align*}
Rather than directly learning $\bmm$ and $\bS$, it is more common to learn the \emph{whitened parameters} \cite{kuss2005assessing,matthews2017scalable}:
$ \bmm' = \bK_{\bZ\bZ}^{- 1/ 2} \bmm$ and $\bS' = \bK_{\bZ\bZ}^{-1 / 2} \bS \bK_{\bZ\bZ}^{-1 / 2}. $
Under these coordinates, the KL divergence term is
$\frac{1}{2} ( \bmm^{\prime \top} \bmm' + \tr{ \bS' } - \log \vert \bS' \vert - M ),$
which doesn't depend on $p(\bu)$ and therefore is relatively simple to optimize.
The posterior distribution $q(f(\bx)) = \normaldist{\ameantest{\bx}}{\avartest{\bx}}$ is given by
\begin{align}
  \ameantest{\bx} = \bk_{\bZ\bx}^\top \bK_{\bZ\bZ}^{-\frac 1 2} \bmm',
  \quad
  \avartest{\bx} = k(\bx, \bx) -
    \bk_{\bZ\bx}^\top \bK_{\bZ\bZ}^{-\frac 1 2} \left( \bI - \bS' \right) \bK_{\bZ\bZ}^{-\frac 1 2} \bk_{\bZ\bx}.
  \label{eqn:approx_pred_dist}
\end{align}

\begin{figure}[t!]
  \centering
  \includegraphics[width=\linewidth]{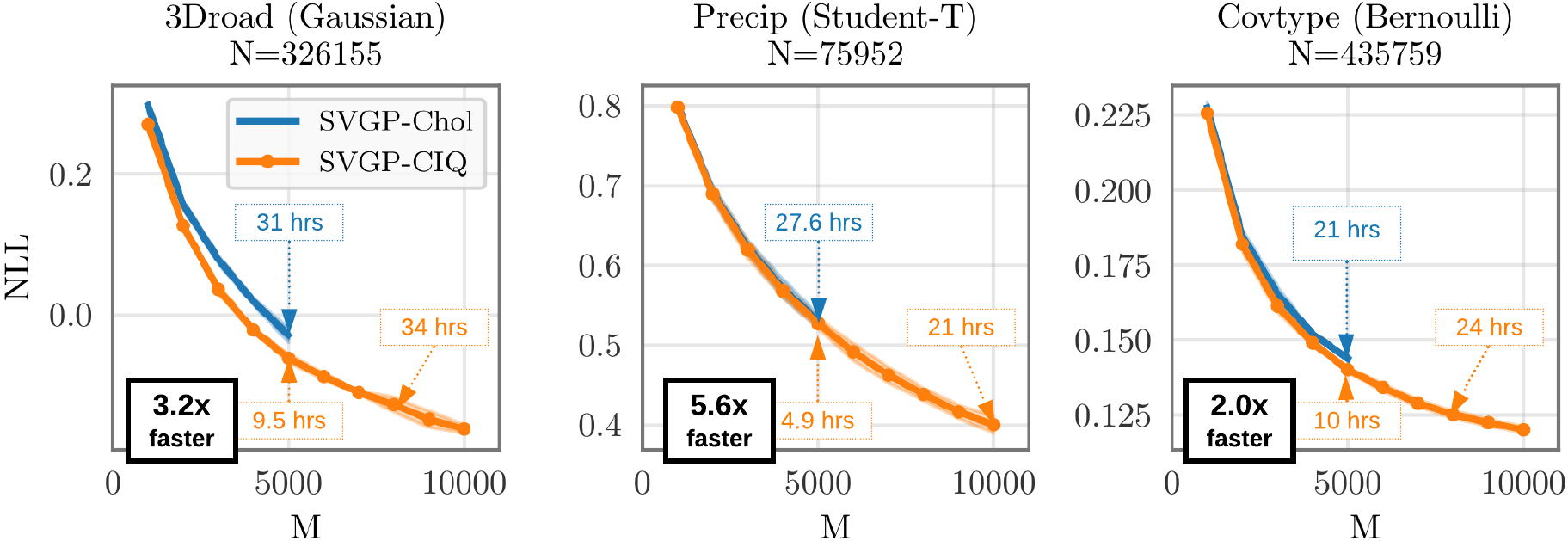}
  \caption[Negative log likelihood (NLL) comparison of Cholesky-whitened vs CIQ-whitened SVGP models.]{
    Negative log likelihood (NLL) comparison of Cholesky vs CIQ SVGP models.
    {\bf Left:} 3DRoad dataset ($N=326155, D=2$, Gaussian likelihood).
    {\bf Middle:} Precipitation dataset ($N=75952, D=3$, Student-T likelihood).
    {\bf Right:} CoverType dataset ($N=435759, D=54$, Bernoulli likelihood).
    NLL improves with more inducing points ($M$), and Cholesky and msMINRES-CIQ models have similar performance.
    However CIQ models train faster than their Cholesky counterparts.
  }
  \label{fig:variational_nll}
\end{figure}

\paragraph{Time and space complexity.}
During training, we repeatedly compute the ELBO and its derivative, which requires computing \cref{eqn:approx_pred_dist} and its derivative for a minibatch of data points.
Optimization typically requires up to $10,\!000$ iterations of training \citep[e.g.][]{salimbeni2018natural}.
We note that $\bK_{\bZ\bZ}^{-1 / 2} \bb$ (and its derivative) is the most expensive numerical operation during each ELBO computation.
If we use Cholesky to compute this operation, the time complexity of SVGP training is $\bigo{M^3}$.\footnote{
  Note that Cholesky computes $\bK^{-1/2} \bb$ up to an orthogonal rotation, which is suitable for whitened SVGP.
}
On the other hand, msMINRES-CIQ-based SVGP training is only $\bigo{J M^2}$, where $J$ is the number of msMINRES iterations.
Both methods require $\bigo{M^2}$ storage for the $\bmm'$ and $\bS'$ parameters.

\paragraph{Natural gradient descent with msMINRES-CIQ.}
The size of the variational parameters $\bmm'$ and $\bS'$ grows quadratically with $M$.
This poses a challenging optimization problem for standard gradient descent methods.
To adapt to the large $M$ regime, we rely on {\bf natural gradient descent (NGD)} to optimize $\bmm'$ and $\bS'$ \citep[e.g.][]{hensman2012fast,salimbeni2018natural}.
At a high level, these methods perform the updates $[\bmm, \:\: \bS] \gets [\bmm, \:\: \bS] - \varphi \: \bFS^{-1} \: \nabla \loglik_\text{ELBO}$,
where $\varphi$ is a step size, $\nabla \loglik_\text{ELBO}$ is the ELBO gradient, and $\bFS$ is the {Fisher information matrix} of the variational parameters.
Na\"ively, each NGD step requires $\bigo{M^3}$ computations with $\bmm'$ and $\bS'$, which would dominate the cost of CIQ-based SVGP.
Fortunately, we can derive a natural gradient update that only relies on matrix solves with $\bS'$, which take $\bigo{J M^2}$ time using preconditioned conjugate gradients.
Therefore, using NGD incurs the same \emph{quadratic} asymptotic complexity as msMINRES-CIQ.
See \cref{app:ngd} for the $\bigo{M^2}$ NGD update equations.

\paragraph{Cholesky vs msMINRES-CIQ.}
We compare msMINRES-CIQ-SVGP against Cholesky-SVGP on 3 large-scale datasets: a GIS dataset ({\bf 3droad}, $D=2$) \citep{guo2012ecomark}, a monthly precipitation dataset ({\bf Precipitation}, $D=3$) \citep{lyon2004strength,lyon2005enso}, and a tree cover dataset ({\bf Covtype}, $D=54$) \citep{blackard1999comparative}.\footnote{
	Details on these datasets (including how to acquire them) are in \cref{app:experimental_details}.
}
Each task has between $N=70,\!000$ and $500,\!000$ training data points.
For 3droad we use a Gaussian observation model.
The Precipitation dataset has noisier observations; therefore we apply a Student-T observation model.
Finally, we reduce the CovType dataset to a binary classification problem and apply a Bernoulli observation model.\footnote{
  The task is predicting whether the primary tree cover at a given location is pine trees or other types of trees.
}
We train models with $ 10^3 \le M \le 10^4$.
See \cref{app:experimental_details} for details.

The two methods achieve very similar test-set negative log likelihood (\cref{fig:variational_nll}). We note that there are small
differences in the optimization dynamics, which is to be expected since $\bK_{\bZ\bZ}^{-1/2} \bk_{\bZ\bx}$ can differ by an orthogonal
transformation when computed with msMINRES-CIQ versus Cholesky.
The key difference is the training time:
with $M=5,\!000$ inducing points, msMINRES-CIQ models are up to \emph{5.6x faster} than Cholesky models (on a Titan RTX GPU).
Moreover, msMINRES-CIQ models with $M=8,\!000$-$10,\!000$ take roughly the same amount of time as $M=5,\!000$ Cholesky models.
This speed is due to the rapid convergence of msMINRES---on average $J=100$ kernel-vector multiplies suffices to achieve 3 decimal places of error (see \cref{app:additional_results}).
Note we do not train $M > 5,\!000$ Cholesky models as doing so would require 14GB of GPU memory and $2$-$10$ days for training.

\paragraph{Effects of increased inducing points.}
We find that accuracy improves with increased $M$ on all datasets.
Scaling from $M=5,\!000$ to $M=10,\!000$ reduces test-set NLL by $0.1$ nats on the 3droad and Precipitation datasets.
We find similar reductions in predictive error (see \cref{app:additional_results} for plots).
By scaling more readily to large $M$, msMINRES-CIQ enables high-fidelity variational approximations that would be computationally prohibitive with Cholesky.

\subsection{Posterior Sampling for Bayesian Optimization}
\label{sec:bayesopt_results}

\begin{figure}[t!]
  \centering
  \includegraphics[width=\linewidth]{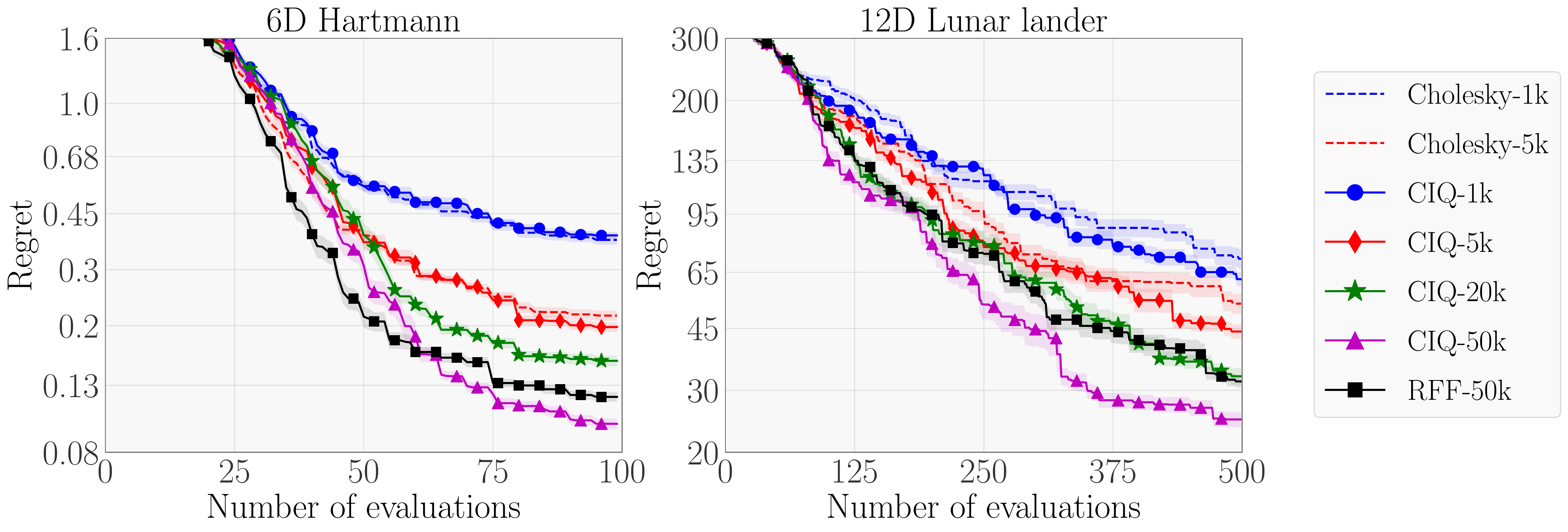}
  \caption[
    A comparison of sampling methods for Bayesian optimization (BO) via TS.
    TS is applied to the Hartmann ($D=6$) and Lunar Lander ($D=12$).
  ]{
    A comparison of sampling methods for Bayesian Optimization.
    BO is applied to the ({\bf left}) Hartmann ($D=6$) and ({\bf right}) Lunar Lander ($D=12$) problems.
    Methods: Cholesky-$\langle T \rangle$ draws posterior samples with Cholesky at $T$ candidate points.
    CIQ-$\langle T \rangle$ draws posterior samples with msMINRES-CIQ.
    RFF-50k uses random Fourier features to draw approximate posterior samples at $50,\!000$ candidate points.
    Larger $T$ results in better optimization.
    msMINRES-CIQ enables scaling to $T \geq 50,\!000$.
    Each plot shows mean regret with standard error in log-scale based on 30 replications.
  }
  \label{fig:bayesopt}
\end{figure}

The second application of msMINRES-CIQ we explore is Gaussian process posterior sampling in the context of Bayesian optimization (BO) \citep[e.g.][]{snoek2012practical}.
Many acquisition functions require drawing samples from posteriors \citep[e.g.][]{frazier2009knowledge,hernandez2014predictive,wang2017max}.
One canonical example is {\bf Thompson Sampling} (TS) \cite{thompson1933likelihood, hernandez2017parallel, kandasamy2018parallelised}.
TS trades off exploitation of existing minima for exploration of new potential minima.
TS chooses the next query point $\widetilde \bx $ as the minimizer of a sample drawn from the posterior.
Let $\bXtest = [ \bxtest_1, \ldots, \bxtest_T ]$ be a \emph{candidate set} of possible query points.
To choose the next query point $\widetilde \bx$, TS computes
\begin{equation}
  \widetilde \bx = \argmin \left( \bmeantest(\bXtest) + {\Covtest(\bXtest)}^{\frac 1 2} \bepsilon \right),
  \quad
  \bepsilon \sim \normaldist{\bzero}{\bI}.
  \label{eqn:thompson_sample}
\end{equation}
where $\bmeantest(\bXtest)$ and $\Covtest(\bXtest)$ are the posterior mean and covariance of the Gaussian process at the candidate set.
The candidate set is often chosen using a space-filling design, e.g.~a Sobol sequence.
The search space grows exponentially with the dimension; therefore, we need large values of $T$ to more densely cover the search space for better optimization performance.
Using Cholesky to compute \cref{eqn:thompson_sample} incurs a $\bigo{T^3}$ computational cost and $\bigo{T^2}$ memory, which severely limits the size of $T$.
In comparison, msMINRES-CIQ only requires $\bigo{T^2}$ computation and $\bigo{T}$ memory.

We perform BO using TS on the classic test function ({\bf Hartmann}, $D=6$) and a reinforcement controller tuning problem ({\bf Lunar Lander}, $D=12$) from the OpenAI gym.\footnote{\url{https://gym.openai.com/envs/LunarLander-v2}}
We provide more details in the supplementary material.
For each problem we use exact Gaussian processes as the surrogate model and TS as the acquisition function.
Our goal is to determine whether CIQ-based sampling is beneficial by enabling scaling to larger candidate set sizes.

\paragraph{Baselines.}
We measure the performance of TS as a function of the candidate set size $T$ and consider $T \in \{ 1,\!000, 5,\!000, 20,\!000, 50,\!000 \}$.
We run Cholesky ({\bf Cholesky}-$T$) for  $T \in \{ 1,\!000, 5,\!000\}$ and msMINRES-CIQ ({\bf CIQ}-$T$) for $T \ge 5,\!000$.
Note that it would be very challenging and impractical to use Cholesky with $T \geq 10,\!000$, due to its quadratic memory and cubic time complexity.
For example, running Cholesky for $T = 50,\!000$ would require $\geq 100$ GB of GPU memory, and performing a single decomposition would take (at best) $\approx 30$ seconds.
In addition to Cholesky and CIQ with exact Gaussian processes as the surrogate model, we also compare to random Fourier features (RFF) \cite{rahimi2008random} with $1,\!000$ random features.

\paragraph{Optimization performance.}
We plot the mean regret with standard error based on 30 replications in \cref{fig:bayesopt}.
By increasing $T=1,\!000$ to $T=50,\!000$, the final regret achieved by CIQ is significantly lower on both problems.
We re-iterate that $T=50,\!000$ is largely impractical with Cholesky.
Large candidate sets have previously only been possible with approximate sampling methods like RFF.
We note, however, that RFF with $T=50,\!000$ is outperformed by CIQ-50k on both problems.

\subsection{Gibbs Samplers and Image Reconstruction}
\label{sec:superres_results}

\begin{figure}[t!]
  \centering
  \includegraphics[width=0.22\linewidth]{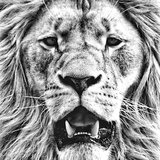} \quad
  \includegraphics[width=0.22\linewidth]{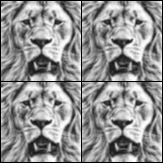} \quad
  \includegraphics[width=0.22\linewidth]{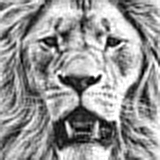} \quad
  \includegraphics[width=0.22\linewidth]{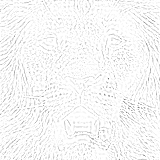}
  \caption{
    Using msMINRES-CIQ for solving problems in spatial statistics, such as image reconstruction.
    This requires sampling from a precision matrix of dimension $D=25,\!600$.
    ({\bf Left}) High-resolution image of dimension $D$.
    ({\bf Middle Left}) Low-resolution images.
    ({\bf Middle Right}) Reconstructed image.
    ({\bf Right}) Delta between original image and reconstruction (darker colors correspond to larger deltas).
  }
  \label{fig:lion}
\end{figure}

High-dimensional Gaussian distributions are ubiquitous in Bayesian statistics, especially in the context of spatially structured data.
Application areas are numerous, including disease mapping, archaeology, and image analysis \cite{besag1991bayesian, waller1997hierarchical, knorr2002block}. Many of the models that arise in these applications are amenable to Gibbs sampling, a MCMC method for generating (approximate) samples
from Bayesian posteriors. As such, sampling from high-dimensional Gaussian distributions is often the primary computational bottleneck for these methods.

To illustrate the utility of msMINRES-CIQ for constructing efficient Gibbs samplers for high-dimensional Gaussian latent variables,
we consider an image reconstruction task \cite{bardsley2012mcmc}. We emphasize,
however, the wide-ranging applicability of these methods, including for non-spatially structured data (e.g. for sparse linear regression \cite{griffin2017hierarchical}).
We formulate an image analysis model as follows: we observe $R$ low-resolution images $\{ \by_r \}_{r=1}^R$, with
each image of size $M \times M$. The goal is to reconstruct the unknown high-resolution image $\bx$ of size $N \times N$ with
$N > M$. The joint density is given by
\begin{equation}
p(\bx, \by_{1:R}, \gamma_{\rm obs}, \gamma_{\rm prior}) = \NN(\by_{1:R} | \bA \bx, \gamma_{\rm obs}^{-1} \bone)
\NN(\bx | \mathbf{0}, \gamma_{\rm prior}^{-1}\bL) p(\gamma_{\rm obs}) p(\gamma_{\rm prior})
\end{equation}
where $\bA$ is a $M^2R \times N^2$ matrix that encodes how the high-resolution image is blurred and down-sampled to yield $R$ low-resolution images and
$\bL$ is a $N^2 \times N^2$ discrete Laplace operator that encodes our prior smoothness assumptions about the image $\bx$. Additionally,
$\gamma_{\rm obs}$ and $\gamma_{\rm prior}$ are scalar hyperparameters that control the scale of the observation noise and strength of the image prior, respectively. For more details please refer to Appendix~\ref{app:experimental_details}.
The computational bottleneck in the resulting Gibbs sampler is sampling from the conditional Gaussian distribution given by
\begin{equation}
\begin{split}
\nonumber
p(\bx | \by_{1:R}, \gamma_{\rm obs}, \gamma_{\rm prior}) = \NN(\bx | \bmm, \bLam^{-1}) \;\;\;\;\;
\bmm = \gamma_{\rm obs} \bLam^{-1} \bA^{T} \by_{1:R} \;\;\;\;\;
\bLam = \gamma_{\rm obs} \bA^{T} \bA + \gamma_{\rm prior} \bL
\end{split}
\end{equation}
For a concrete demonstration we perform image reconstruction on the image depicted in \cref{fig:lion}. Here
$N=160$, $M=80$, and $R=4$, so that the precision matrix $\bLam$ is of size $25600 \times 25600$. Despite the
extreme size, our implementation achieves $\approx 0.61$ samples per second (using a TitanRTX GPU).
\new{
We estimate that a Cholesky version of this method would achieve only $\approx 0.05$ samples per second.
}

\section{Discussion}

We have introduced msMINRES-CIQ---a MVM-based method for computing $\bK^{1/2} \bb$ and $\bK^{-1/2} \bb$.
In sampling and whitening applications, msMINRES-CIQ can be used as a $\bigo{N^2}$ drop-in replacement for the $\bigo{N^3}$ Cholesky decomposition.
Its scalability and GPU utilization enable us to use more inducing points with SVGP models and larger candidate sets in Bayesian optimization.
In all applications, such increased fidelity results in better performance.

\new{
\paragraph{Stability of msMINRES-CIQ.}
Krylov methods on symmetric matrices can be prone to numerical instabilities due to round-off errors \citep[e.g.][]{parlett1979lanczos}.
Our method has two key advantages that improve stability.
First, we only use Krylov methods to solve linear systems rather than eigenvalue problems.
Common numerical pitfalls that hinder Krylov eigen-solvers (e.g.~loss of orthogonality between Lanczos vectors) have been shown to have little empirical effect on linear system solvers like MINRES and CG \citep[e.g.][]{trefethen1997numerical,fong2012cg}.
Second, each solve from msMINRES is inherently a shifted system $\bK + t_q \bI$.
In practice these shifts dramatically improve the conditioning of $\bK$, and allow us to work directly with the matrix $\bK$ without having to add diagonal jitter for stability.

\paragraph{Comparison to other fast sampling methods.}
Historically, GP samples have been drawn using the Cholesky factor or finite-basis approximations like RFFs.
Recently, a growing line of work investigates using inducing point methods for scalable sampling \cite{pleiss2018constant,wilson2020efficiently}.
We believe that CIQ-sampling can be used in conjunction with these inducing point approaches.
For example, \citet{wilson2020efficiently} use RFFs to sample from the prior and an inducing point approximation of the conditional to convert prior samples into posterior samples.
CIQ can augment this approach, allowing for more inducing points and/or replacing RFFs for prior sampling.
}

\paragraph{Advantages and disadvantages.}
One advantage of the Cholesky decomposition is its reusability.
As discussed in \cref{sec:empirical}, the cubic cost of computing $\bL \bL^\top$ is amortized when drawing $\bigo{M}$ samples or whitening $\bigo{M}$ vectors.
Conversely, applying msMINRES-CIQ to $\bigo{M}$ vectors would incur a $\bigo{M^3}$ cost, eroding its computational benefits. Thus, our method is primarily advantageous
in scenarios with a small number of right hand sides or where $\bK$ is too large to apply Cholesky.
We also emphasize that msMINRES-CIQ---like all Krylov methods---can take advantage of fast MVMs afforded by structured covariances.
Though this paper focuses on applying this algorithm to dense matrices, we suggest that future work explore applications involving sparse or structured matrices.

\section*{Broader Impact}

This paper introduces an algorithm to improve the efficiency and scalability of a common-place computation.
The results section highlights three common use cases of this algorithm: variational Gaussian processes, Bayesian optimization, and Gibbs sampling.
While there are other potential use-cases of this method, we will focus on the broader impacts with respect to these three applications.

Variational Gaussian processes and Gibbs sampling are common methods.
Other researchers have focused on domains like medicine \citep{schulam2015framework,futoma2017learning}, geo-statistics \citep{diggle1998model,stein2012interpolation}, and time-series modelling \citep{roberts2013gaussian,wilson2013gaussian} to motivate the need for increased scalability and efficiency.
We believe that our proposed algorithm will make Gaussian process models and Gibbs sampling techniques increasingly applicable in these settings.
Researchers/practitioners in these fields might have previously been unable to use Gaussian processes/Gibbs sampling due to scalability issues.
While we believe increasing the scalability and usability of these probabilistic techniques is a worthwhile goal, we note that they require additional care when using.
If a system is to rely on probabilistic methods for calibrated uncertainty estimates, it will no longer be sufficient to iterate on accuracy as a target method.
We also note that performing meaningful probabilistic inferences requires some level of domain expertise regarding modeling priors and potential biases of sampling/variational approximations.

Bayesian optimization is a tool commonly used for hyperparameter optimization \cite{snoek2012practical}, A/B testing \cite{balandat2019botorch}, and other black-box optimization problems.
One of the most popular and best performing acquisition functions is Thompson sampling, which requires sampling the unknown function at a candidate set.
The primary benefit of the proposed method is better optimization, which could lead to better machine learning models (via better hyperparameter searches) and faster experimental testing (via A/B testing).
We would argue that improving the efficiency of such algorithms poses minimal risk beyond more general concerns about potential misapplications of the underlying technology to the optimization of nefarious objectives, intentionally or otherwise.
However, we will make note here of some general risks associated with black-box optimization: a potential over-reliance on fully automated methods and computationally expensive searches for what might be marginal improvements.

We have release an open-sourced implementation of this algorithm to facilitate the adoption of this method.\footnote{See \url{bit.ly/ciq_svgp} and \url{bit.ly/ciq_sampling}.}
Since our method relies on quadrature approximations and iterative refinement, one mode of failure is when such iterations fail to converge to a good estimate (for example, due to bad conditioning).
However, there are several easy-to-perform convergence checks (e.g. the msMINRES residual), and such convergence checks are part of our implementation to catch such failure cases.

\begin{ack}
  We thank David Bindel for helpful conversations about rational approximations and optimization.
  At the time of submission, GP was support by grants from the National Science Foundation NSF (III-1618134, III-1526012, IIS-1149882, IIS- 1724282, OAC-1934714, and TRIPODS-1740822), the Office of Naval Research DOD (N00014-17-1-2175), the Bill and Melinda Gates Foundation, and the Cornell Center for Materials Research with funding from the NSF MRSEC program (DMR-1719875).
  AD is partially funded by the National Science Foundation under award DMS-1830274
  We are thankful for generous support by Zillow and SAP America Inc.
\end{ack}

\addcontentsline{toc}{section}{References}
{\small
  \bibliographystyle{abbrvnat}
  \bibliography{citations}
}

%
%

\clearpage


\makeatletter
  \setcounter{table}{0}
  \renewcommand{\thetable}{S\arabic{table}}%
  \setcounter{figure}{0}
  \renewcommand{\thefigure}{S\arabic{figure}}%
  \setcounter{equation}{0}
  \renewcommand\theequation{S\arabic{equation}}
  \renewcommand{\bibnumfmt}[1]{[S#1]}

  \newcommand{\suptitle}{Supplementary Information for: \titl}
  \renewcommand{\@title}{\suptitle}
  \newcommand{\thanks}[1]{\footnotemark[1]}
  \renewcommand{\@author}{\authorinfo}

  \par
  \begingroup
    \renewcommand{\thefootnote}{\fnsymbol{footnote}}
    \renewcommand{\@makefnmark}{\hbox to \z@{$^{\@thefnmark}$\hss}}
    \renewcommand{\@makefntext}[1]{%
      \parindent 1em\noindent
      \hbox to 1.8em{\hss $\m@th ^{\@thefnmark}$}#1
    }
    \thispagestyle{empty}
    \@maketitle
    \@thanks
  \endgroup
  \let\maketitle\relax
  \let\thanks\relax
\makeatother

\appendix

\begin{algorithm2e}[h!]
  \SetAlgoLined
  \SetKwInOut{Input}{Input}
  \SetKwInOut{Output}{Output}
  \newcommand\NextInput[1]{%
    \settowidth\inputlen{\Input{}}%
    \setlength\hangindent{1.5\inputlen}%
    \hspace*{\inputlen}#1\\
  }
  \newcommand\graycomment[1]{\footnotesize\ttfamily\textcolor{gray}{#1}}
  \SetCommentSty{graycomment}
  \SetKw{Break}{break}
  \SetKwFunction{mvmkxx}{mvm\_$\bK$}
  \SetKwFunction{computequad}{compute\_quad}
  \SetKwFunction{lanczos}{lanczos}
  \SetKwFunction{minres}{msMINRES}
  \SetKwFunction{size}{size}
  \caption{Computing $\bK^{-\frac 1 2} \bb$ with MVM-based Contour Integral Quadrature (CIQ)}
  \label{alg:ciq}
    \Input{\mvmkxx{$\cdot$} -- function for matrix-vector multiplication (MVM) with matrix $\bK$}
    \NextInput{$\bb$ -- right hand side, $J$ -- number of \minres iterations, $Q$ -- number of quad. points}
    \Output{$\ba \approx \bK^{- \frac 1 2} \bb$}
    \BlankLine
    $[w_1, \ldots, w_Q]$, $[t_1, \ldots, t_Q]$ $\gets$ \computequad{ \mvmkxx{$\cdot$}, $Q$}
    \tcp{Weights ($w_i$) and shifts ($t_i$) for quadrature -- details in \cref{app:quadrature}.}
    $(t_1 \bI + \bK)^{-1}$\bb, $\ldots$ $(t_Q \bI + \bK)^{-1}\bb$ $\gets$ \minres{ \mvmkxx{$\cdot$}, $\bb$, $J$, $t_1$, $\ldots$, $t_Q$}
    \tcp{msMINRES computes all solves simultaneously -- details in \cref{app:minres}.}
    \BlankLine
    \Return{$\sum_{q=1}^Q w_q \left( t_q \bI + \bK \right)^{-1} \bb$}
    \tcp{CIQ estimate of $\tfrac{1}{2\pi i} \int \tau^{-1 / 2} (\tau \bI - \bK)^{-1} \bb \intd \tau = \bK^{-1 / 2} \bb$}
\end{algorithm2e}

\section{Additional Results}
\label{app:additional_results}

\begin{figure}[ht!]
	\centering
	\includegraphics[width=\textwidth]{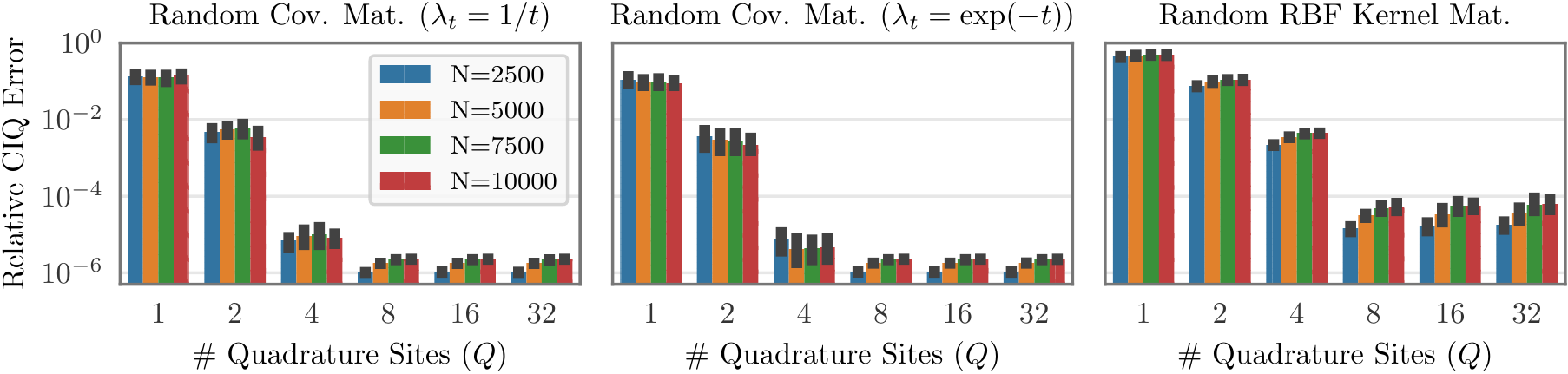}
  \caption{
    CIQ relative error at computing $\bK^{1/2} \bb$ as a function of number of quadrature points $Q$.
    In all cases $Q=8$ achieves $<10^{-4}$ error.
  }
  \label{fig:quad_error_supp}
\end{figure}

\begin{figure}[ht!]
	\centering
	\includegraphics[width=\textwidth]{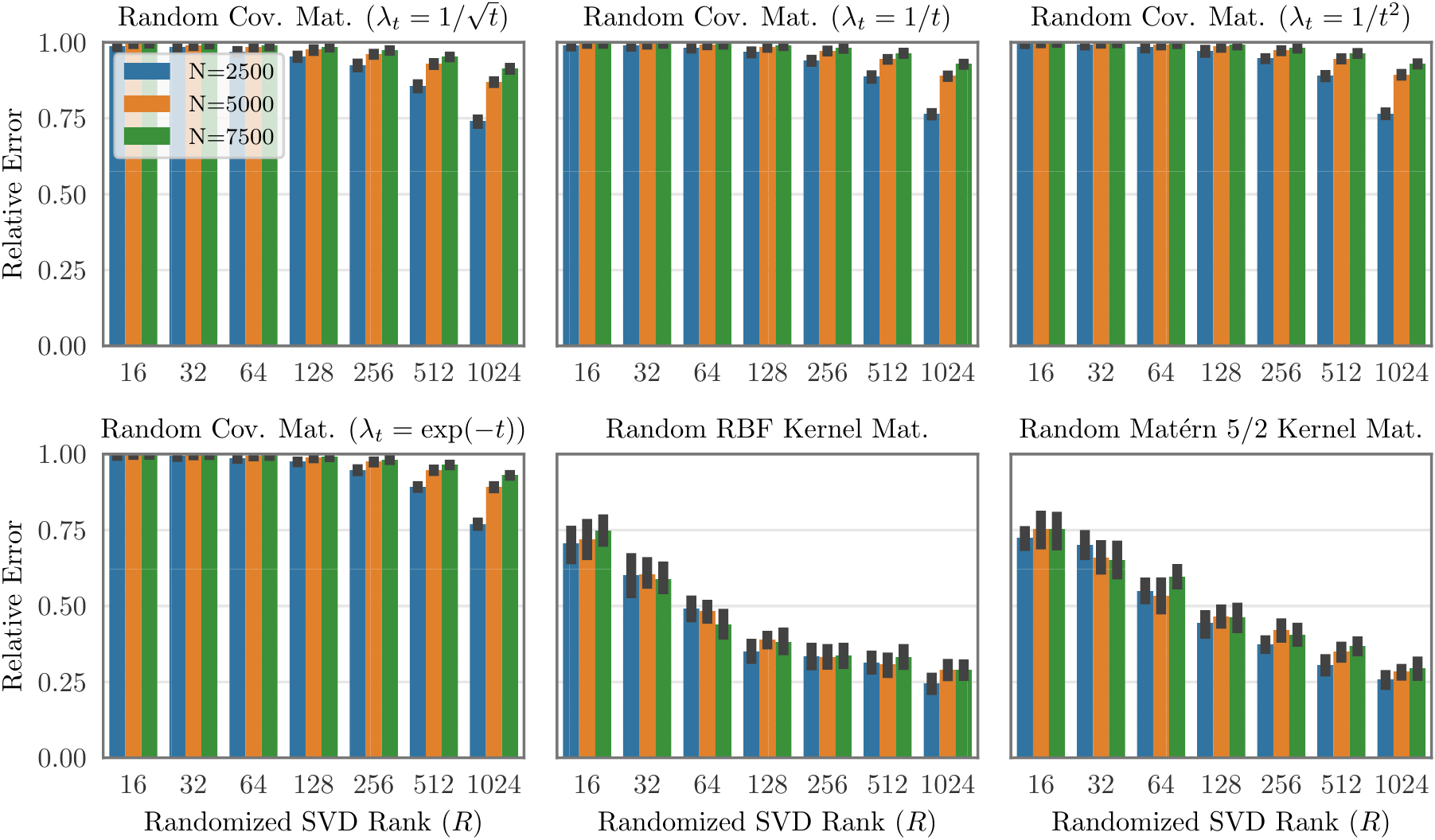}
  \caption{
    Randomized SVD relative error at computing $\bK^{1/2} \bb$ as a function of approximation rank $R$.
    In all cases, randomized SVD is unable to achieve a relative error better than about $0.25$.
  }
  \label{fig:randomized_svd_supp}
\end{figure}

\cref{fig:quad_error_supp} and \cref{fig:randomized_svd_supp} are continuations of \cref{fig:quad_error}.
They plots CIQ convergence and randomized SVD convergence as a function of $Q$ and $R$ for covariance matrices whose eigenvalues decay as $\lambda_{t}=\frac{1}{\sqrt{t}}$, $\lambda_{t}=\frac{1}{t}$, $\lambda_{t}=\frac{1}{t^2}$, and $\lambda_{t}=\exp(-t)$ in addition
to the kernel matrix results already presented. The results for CIQ demonstrate that it is relatively invariant to the eigenvalue decay speed, and does not require approximately low rank structure.
Randomized SVD on the other hand incurs an order of magnitude more error; a rank of $1,\!024$ is unable to reduce the relative error to a single decimal point.

\begin{figure}[ht!]
	\centering
	\includegraphics[width=0.7\textwidth]{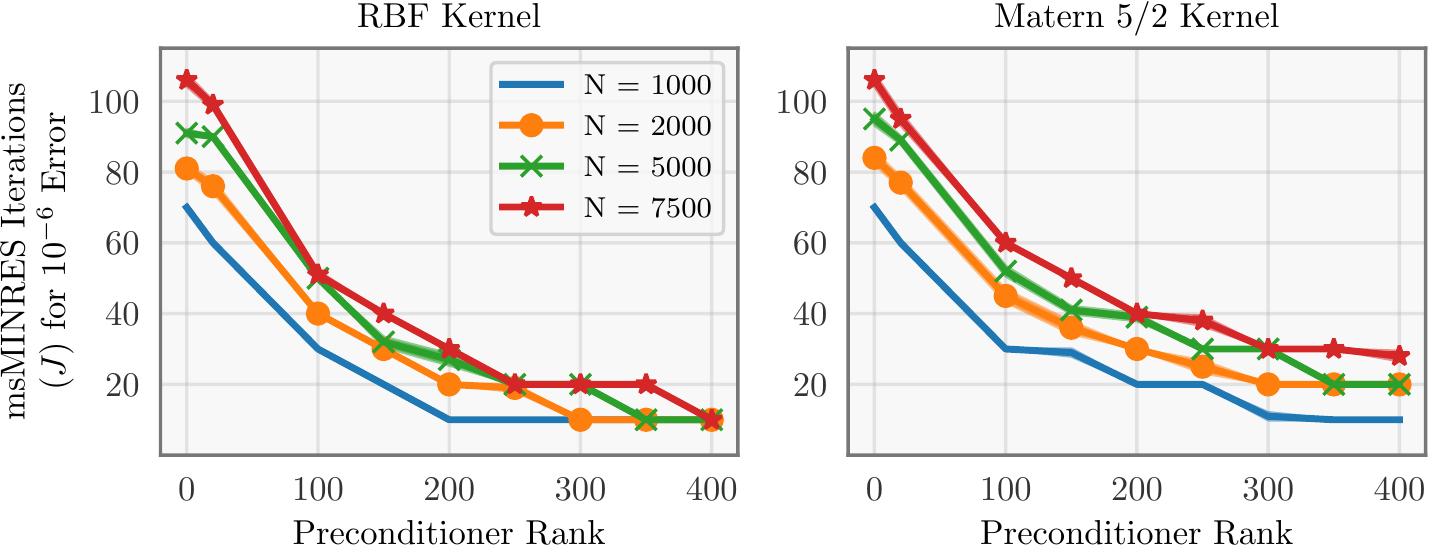}
  \caption{
    Effect of preconditioning on CIQ convergence (random RBF and Mat\'ern-5/2 kernels with a pivoted Cholesky preconditioner \citep{gardner2018gpytorch}).
  }
  \label{fig:precond_result_rbf}
\end{figure}

\cref{fig:precond_result_rbf} further demonstrates the effect of preconditioning on msMINRES-CIQ.
We construct random $N \times N$ RBF/Mat\'ern kernels, applying msMINRES-CIQ to a set of $N$ orthonormal vectors ($[\bK^{1/2} \bb_{1}, \ldots, \bK^{1/2} \bb_{N}]$), and compute the empirical covariance.
We plot the number of msMINRES iterations needed to achieve a relative error of $10^{-4}$.
The pivoted Cholesky preconditioner of \citet{gardner2018gpytorch}---which forms a low-rank approximation of $\bK$---accelerates convergence of msMINRES.
Without preconditioning (i.e.~rank=0), $J=100$ iterations are required for $N=7,\!500$ matrices.
With rank-100/rank-400 preconditioners, iterations are cut by a factor of two/four.

\begin{figure}[ht!]
	\centering
	\includegraphics[width=0.7\textwidth]{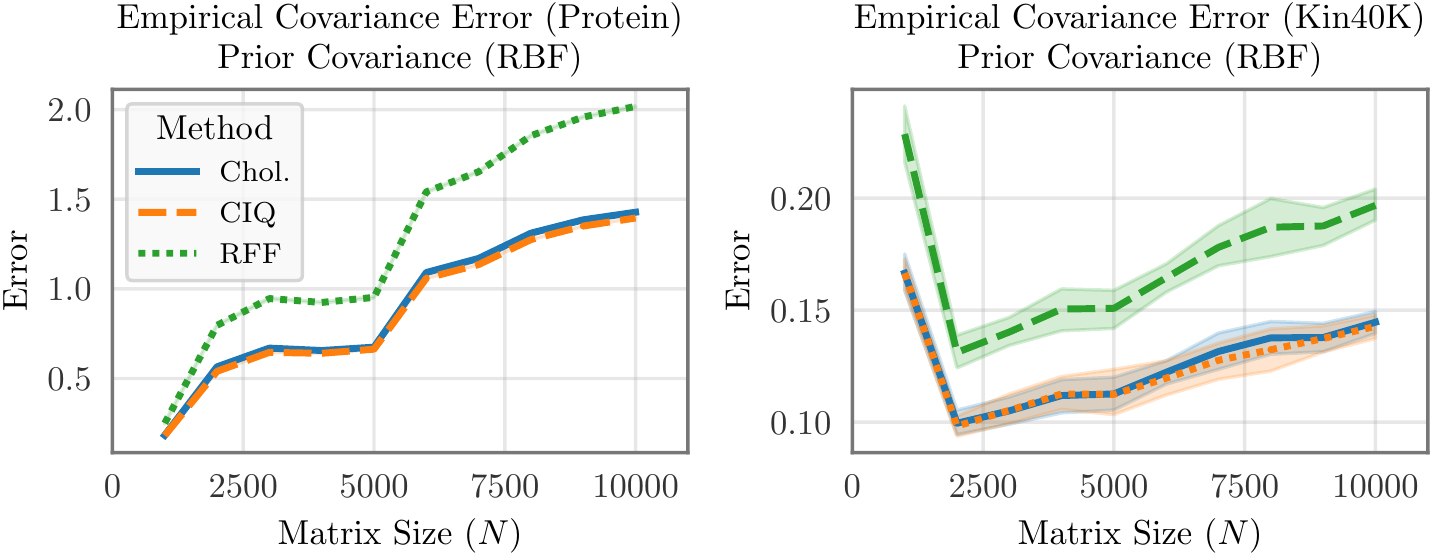}
  \caption{
    Empirical covariance error (relative norm) for various sampling methods (Cholesky, msMINRES-CIQ, and $1,\!000$ Random Fourier Features \cite{rahimi2008random}).
    Empirical covariances are measured from $1,\!000$ samples.
    RBF matrices are constructed from data in the Protein and Kin40k datasets \cite{asuncion2007uci}.
  }
  \label{fig:empirical_covariance_matrix}
\end{figure}

To further compare msMINRES-CIQ to randomized methods, \cref{fig:empirical_covariance_matrix} plots the empirical covariance matrix of $1,\!000$ Gaussian samples drawn from a Gaussian process prior $\normaldist{\bzero}{\bK}$.
We construct the RBF covariance matrices $\bK$ using subsets of the Protein and Kin40k datasets\footnote{
  Both datasets are originally from the UCI repository and can be downloaded from
  \url{https://github.com/gpleiss/ciq_experiments/tree/main/svgp/data}.
} \cite{asuncion2007uci}.
We note that all methods incur some sampling error, regardless of the subset size ($N$).
msMINRES-CIQ and Cholesky-based sampling tend to have very similar empirical covariance error.
On the other hand, the Random Fourier Features method \cite{rahimi2008random} (with $1,\!000$ random features) incurs errors up to $2\times$ as large.
This additional error is due to the randomness in the RFF approximation.

\begin{figure}[ht!]
  \centering
  \includegraphics[width=\linewidth]{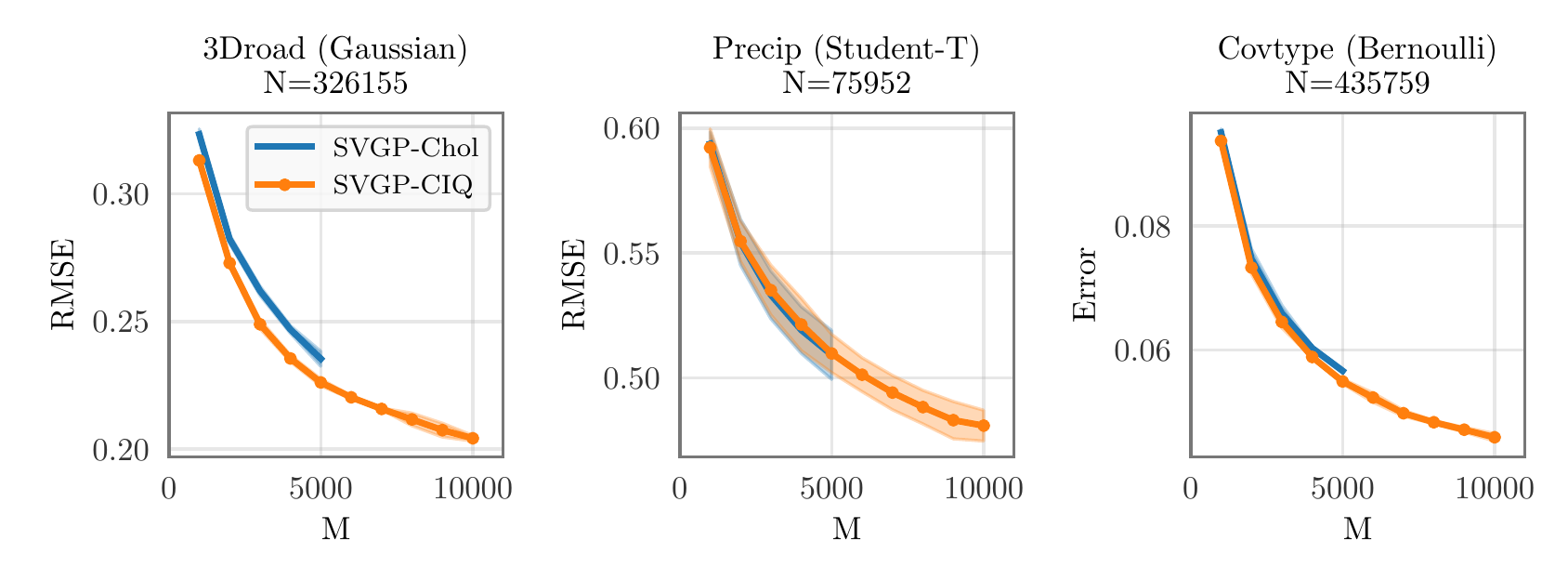}
  \caption[Error comparison of Cholesky-whitened vs CIQ-whitened SVGP models.]{
    Error comparison of Cholesky-whitened vs CIQ-whitened SVGP models.
    {\bf Left:} 3DRoad dataset RMSE ($N=326155, D=2$, Gaussian likelihood).
    {\bf Middle:} Precipitation dataset RMSE ($N=75952, D=3$, Student-T likelihood).
    {\bf Right:} CoverType dataset $0/1$ error ($N=435759, D=54$, Bernoulli likelihood).
    Error improves with more inducing points ($M$), and Cholesky and CIQ models have similar performance.
    However CIQ scales to larger values of $M$.
  }
  \label{fig:variational_error}
\end{figure}

In \cref{fig:variational_error} we plot the predictive error of CIQ-SVGP and Chol-SVGP models as a function of $M$.
For the two regression datasets (3droad and Precipitation) error is measured by test set root mean squared error (RMSE).
On the Covtype classification dataset error is measured by the test set $0/1$ loss.
As with the NLL results in \cref{fig:variational_nll} we find that the CIQ-SVGP and Chol-SVGP perform similarly, despite the fact that CIQ-SVGP can be up to $5.6\times$ faster.
Moreover, we see that error continuously decreases with more inducing points up to $M=10,\!000$.

\begin{figure}[ht!]
  \centering
  \includegraphics[width=\linewidth]{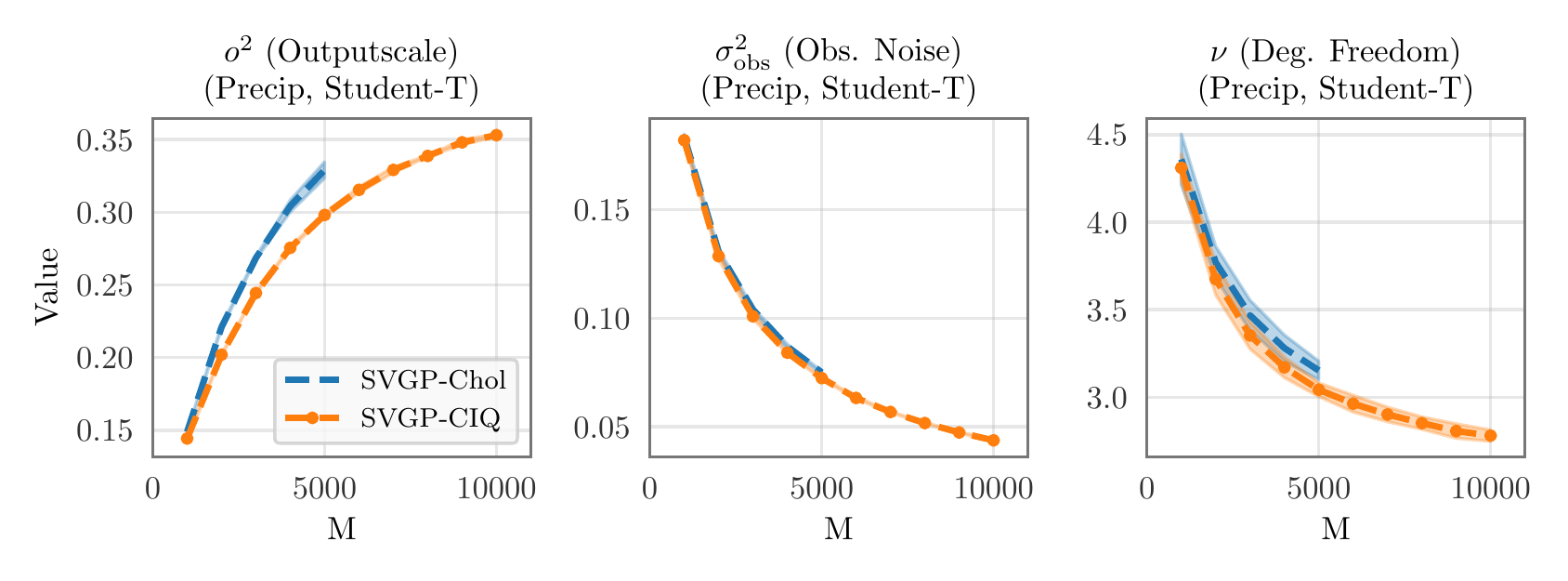}
  \caption{
    Hyperparameters versus number of inducing points ($M$) for Chol-SVGP and CIQ-SVGP (Precipitation dataset, Student-T likelihood).
    As $M$ increases, the kernel outputscale (left) also increases.
    At the same time, the estimated observational noise (middle) decreases as does the estimated degrees of freedom (right), reflecting a heavier-tailed noise distribution.
    This suggests that, with larger $M$, SVGP models can find more signal in the data.
  }
  \label{fig:variational_stats}
\end{figure}

In \cref{fig:variational_stats} we plot the learned hyperparameters of the Precipitation SVGP models:
\begin{enumerate*}
  \item $o^2$ (the kernel outputscale)---which roughly corresponds to variance explained as ``signal'' in the data;
  \item $\sigma^2_\text{obs}$---which roughly corresponds to variance explained away as observational noise; and
  \item $\nu$ (degrees of freedom)---which controls the tails of the noise model (lower $\nu$ corresponds to heavier tails).
\end{enumerate*}
As $M$ increases, we find that the observational noise parameter decreases by a factor of $4$---down from $0.19$ to $0.05$---while the $\nu$ parameter also decreases.
Models with larger $M$ values can more closely approximate the true posterior \cite{hensman2013gaussian}; therefore, we expect that the parameters from the larger-$M$ likelihoods more closely correspond to the true dataset noise.
This confirms findings from \citet{bauer2016understanding}, who argue that variational approximations with small $M$ can tend to overestimate the amount of noise in datasets.

\new{
\begin{figure}[ht!]
  \centering
  \includegraphics[width=0.66\linewidth]{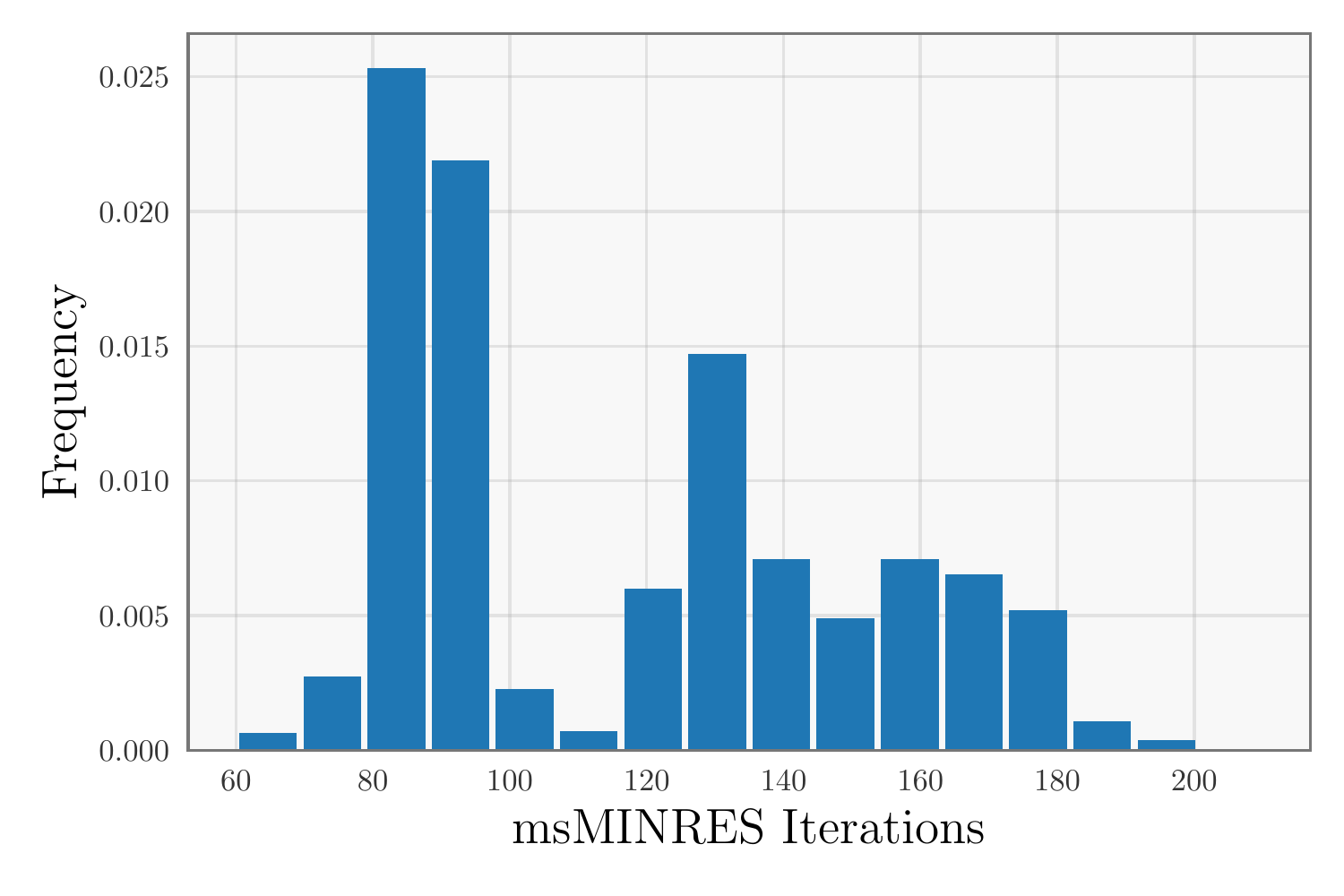}
  \caption{
    Number of msMINRES iterations needed to achieve a relative residual of $10^{-3}$.
    Histogram captures training a $M=5,\!000$ SVGP model on the 3droad dataset (subsampled to $30,\!000$ data points).
  }
  \label{fig:msminres_iters}
\end{figure}

\cref{fig:msminres_iters} is a histogram displaying the msMINRES iterations needed to achieve a relative residual of $10^{-3}$ when training a $M=5,\!000$ SVGP model on the 3droad dataset (subsampled to $30,\!000$ data points).
Most msMINRES calls converge in fewer than $100$ iterations; almost no calls require more than $200$ iterations.
We hypothesize that this fast convergence is due to solving shifted systems ($\bK + t_q \bI$).
The minimum eigenvalues of the shifted matrix are lower-bounded by $t_q$, and therefore shifted systems have a better condition number than the unshifted matrix $\bK$.
}

\section{Quadrature for Matrix Square Roots}
\label{app:quadrature}

Here we briefly describe the quadrature formula derived by \citet{hale2008computing} for use with Cauchy's integral formula and refer the reader to the original publication for more details.

Assume that $\bK$ is a positive definite matrix, and thus has real positive eigenvalues.
Our goal is to approximate Cauchy's integral formula with a quadrature estimate:
\begin{align}
	f(\bK)
  &= \frac{1}{2 \pi i} \oint_\Gamma f(\tau) \left( \tau \bI - \bK \right)^{-1} \intd \tau
  \label{eqn:contour_integral_2}
  \\
  &\approx
  \frac{1}{2 \pi i} \sum_{q=1}^Q \widetilde w_q f(\tau_q) \left( \tau_q \bI - \bK \right)^{-1},
  \label{eqn:contour_integral_quad_2}
\end{align}
where $f(\cdot)$ is analytic on and within $\Gamma$, and $\widetilde w_q$ and $\tau_q$ are quadrature weights and nodes respectively.
Note that \cref{eqn:contour_integral_2} holds true for any closed contour $\Gamma$ in the complex plane that winds once (counterclockwise) around the spectrum of $\bK$.

\paragraph{A na\"ive approach with uniformly-spaced quadrature.}
For now, assume that $\lambda_\text{min}$ and $\lambda_\text{max}$---the minimum and maximum eigenvalues of $\bK$---are known.
(We will later address how they can be efficiently estimated.)
A na\"ive first approach to \cref{eqn:contour_integral_quad_2} is to uniformly place the quadrature locations in a circle that surrounds the eigenvalues and avoids crossing the negative real axis, where we anticipate $f$ may be singular:
\[
  \tau_q = \frac{\lambda_{\max}+\lambda_{\min}}{2} + \frac{\lambda_{\max}}{2}e^{2 i \pi \left( q / Q \right)},
  \quad
  \widetilde w_q = \frac 1 Q, \quad q=0,1,\ldots,Q-1.
\]
This corresponds to a standard trapezoid quadrature rule. However, \citet{hale2008computing} demonstrate that the convergence of this quadrature rule depends linearly on the condition number $\kappa(\bK) = \lambda_\text{max} / \lambda_\text{min}$. In particular, this is because the integrand is only analytic in a narrow region around the chosen contour. As many kernel matrices tend to be approximately low-rank and therefore ill-conditioned, this simple quadrature rule requires large $Q$ to achieve the desired numerical accuracy.

\paragraph{Improving convergence with conformal mappings.}
Rather than uniformly spacing the quadrature points, it makes more sense to place more quadrature points near $\lambda_\text{min}$ and fewer near $\lambda_\text{max}$.
This can be accomplished by using the above trapezoid quadrature rule in a {transformed parameter space} that is ``stretched'' near $\lambda_\text{min}$ and contracted near $\lambda_\text{max}$. Mathematically, this is accomplished by applying a conformal mapping that moves the singularities to the upper and lower boundaries of a periodic rectangle. We may then apply the trapezoid rule along a contour traversing the middle of the rectangle---maximizing the region in which the function we are integrating is analytic around the contour.

\subsection{A Specific Quadrature Formula for $f(\bK) = \bK^{-1/2}$}
\citet{hale2008computing} suggest performing a change of variables that projects \cref{eqn:contour_integral_2} onto an annulus.
Uniformly spaced quadrature points inside the annulus will cluster near $\lambda_\text{min}$ when projected back into the complex plane.
This change of variables has a simple analytic formula involving Jacobi elliptic functions (see \citep[][Sec. 2]{hale2008computing} for details.)
In the special case of $f(\bK) = \bK^{-1/2}$, we can utilize an additional change of variables for an even more efficient quadrature formulation \citep[][Sec. 4]{hale2008computing}.
Setting $\sigma = \tau^{1/2}$, we have
\begin{align}
	\bK^{-\frac 1 2}
  &= \frac{1}{\pi i} \oint_{\Gamma_s} \left( \sigma^2 \bI - \bK \right)^{-1} \intd \sigma.
  \nonumber
  \\
  &\approx
  \frac{1}{\pi i} \sum_{q=1}^Q \widetilde w_q \left( \sigma_q^2 \bI - \bK \right)^{-1},
  \label{eqn:contour_integral_quad_3}
\end{align}
where $\Gamma_\sigma$ is a contour that surrounds the spectrum of $\bK^{1/2}$.
Since the integrand is symmetric with respect to the real axis, we only need to consider the imaginary portion of $\Gamma_\sigma$.
Consequently, all the $\tau_q$ quadrature locations (back in the original space) will be real-valued and negative.
Combining this square-root change-of-variables with the annulus change-of-variables results in the following quadrature weights/locations:
\begin{equation}
  \begin{split}
    \sigma_q^2
    &= \lambda_\text{min} \Bigl( \text{sn}(i u_q \mathcal{K}'(k) \mid k) \Bigr)^2,
    \\
    \widetilde w_q
    &= -\frac{ 2 \sqrt{\lambda_\text{min}} }{ \pi Q }
    \:\: \left[
    \mathcal{K}'( k )
    \:\:\: \text{cn} \left( i u_q \mathcal{K}'(k) \mid k \right)
    \:\:\: \text{dn} \left( i u_q \mathcal{K}'(k) \mid k \right)
    \right],
  \end{split}
  \label{eqn:quad_points_and_locations}
\end{equation}
where we adopt the following notation:
\begin{itemize}
  \item $k = \sqrt{ \lambda_\text{min} / \lambda_\text{max} } = 1 / \sqrt{ \kappa(\bK) }$;
  \item $\mathcal{K}'(k)$ is the complete elliptic integral of the first kind with respect to the complimentary elliptic modulus $k' = \sqrt{1 - k^2}$;
  \item $u_q = \frac{1}{Q}(q - \frac 1 2)$; and
  \item $\text{sn}(\cdot \mid k)$, $\text{cn}(\cdot \mid k )$, and $\text{dn}(\cdot \mid k)$ are the Jacobi elliptic functions with respect to elliptic modulus $k$.
\end{itemize}
The weights $\widetilde w_q$ and locations $\sigma_q^2$ from \cref{eqn:quad_points_and_locations} happen to be real-valued and negative.
Setting $t_q = -\sigma_q^2$ and $w_q = -\widetilde w_q$ gives us:
\begin{equation}
	\bK^{-\frac 1 2} \approx \sum_{q=1}^Q w_q \left( t_q \bI + \bK \right)^{-1}, \quad w_q = -\widetilde w_q > 0, \quad t_q = -\sigma_q^2 > 0.
  \label{eqn:contour_integral_quad_4}
\end{equation}
An immediate consequence of this is that the shifted matrices $(t_q \bI + \bK)$ are all positive definite.

\paragraph{Convergence of the quadrature approximation.}
Due to the double change-of-variables, the convergence of this quadrature rule in \cref{eqn:quad_points_and_locations} is extremely rapid---even for ill-conditioned matrices.
\citeauthor{hale2008computing} prove the following error bound:
\begin{lemma}[\citet{hale2008computing}, Thm. 4.1]
  Let $t_1$, $\ldots$, $t_Q > 0$ and $w_1$, $\ldots$, $w_Q > 0$ be the locations and weights of \citeauthor{hale2008computing}'s quadrature procedure.
  The error of \cref{eqn:contour_integral_quad} is bounded by:
  \[
    \left\Vert \bK \sum_{q=1}^Q w_q \left( t_q \bI + \bK \right)^{-1} - \bK^{\frac 1 2} \right\Vert_2
    \leq \bigo{\exp\left( -\frac  {2 Q \pi^2}{\log \kappa(\bK) + 3} \right)},
  \]
  where $\kappa(\bK) = \lambda_\text{max} / \lambda_\text{min}$ is the condition number of $\bK$.
\label{lemma:hale}
\end{lemma}
Remarkably, the error of \cref{eqn:contour_integral_quad} is \emph{logarithmically} dependent on the conditioning of $\bK$.
Consequently, $Q\approx8$ quadrature points is even sufficient for ill-conditioned matrices (e.g. $\kappa(\bK) \approx 10^4$).

\subsection{Estimating the Minimum and Maximum Eigenvalues}
The equations for the quadrature weights/locations depend on the extreme eigenvalues $\lambda_\text{max}$ and $\lambda_\text{min}$ of $\bK$.
Using the Lanczos algorithm \cite{lanczos1950iteration}---which is a Krylov subspace method---we can obtain accurate estimates of these extreme eigenvalues using relatively few matrix-vector multiplies with $\bK$.

\paragraph{The Lanczos algorithm}
is a method for computing an orthonormal basis for Krylov subspaces of a symmetric matrix $\bK$ and, simultaneously, projections of $A$ onto that subspace.
Given an initial vector $\bb$, the algorithm iteratively factorizes $\bK$ as:
\[
  \bK \bQ_J = \bQ_J \bT_J + \br_J \be_J^\top
\]
where $\be_J$ is a unit vector, and
\begin{itemize}
  \item $\bQ_J \in \reals^{N \times J}$ is an orthonormal basis of the $J^\text{th}$ Krylov subspace $\mathcal{K}(\bK, \bb)$,
	\item $\bT_J \in \reals^{J \times J}$ is a symmetric tridiagonal matrix, and
	\item $\br_J \in \reals^J$ is a residual term.
\end{itemize}
At a high level, the Lanczos iterations form the Krylov subspaces while simultaneously performing a process akin to modified Gram Schmidt orthogonalization:
\[
  \text{span} \{ \bq^{(1)}, \:\: \ldots, \:\: \bq^{(J)} \} = \mathcal{K}(\bK, \bb) = \text{span} \{ \bb, \:\: \bK\bb, \:\: \bK^2\bb, \:\: \ldots, \:\: \bK^{J-1} \bb \}.
\]
The orthogonal basis vectors are collected into $\bQ$ and the orthogonalization coefficients are collected into $\bT$.
Due to the symmetry of $\bK$ a three term recurrence exists for this process and each vector $\bq^{(j)}$ only has to be orthogonalized against the two previous basis vectors $\bq^{(j-1)}$, $\bq^{(j-2)}$---resulting in a tridiagonal $\bT$.

\paragraph{Estimating Extreme Eigenvalues from Lanczos.}
To estimate $\lambda_\text{min}$ and $\lambda_\text{max}$ from Lanczos, we perform an eigendecomposition of $\bT_J$.
If $J$ is small (i.e. $J \approx 10$) then this eigendecomposition requires minimal computational resources. In fact, as $\bT_J$ is tridiagonal invoking standard routines allows computation of all the eigenvalues in $\bigo{J^2}$ time.
A well-known convergence result of the Lanczos algorithm is that the extreme eigenvalues of $\bT_J$ tend to converge rapidly to $\lambda_\text{min}$ and $\lambda_\text{max}$ \citep[e.g.][]{saad2003iterative,golub2012matrix}.
Since the Lanczos algorithm always produces underestimates of the largest eigenavlue and overestimates of the smallest it is reasonable to use slightly larger and smaller values in the construction of the quadrature scheme---as we see in Lemma~\ref{lemma:hale}, the necessary number of quadrature nodes is insensitive to small overestimates of the condition number.

\subsection{The Complete Quadrature Algorithm}
\cref{alg:quadrature} obtains the quadrature weights $w_q$ and locations $t_q$ corresponding to \cref{eqn:quad_points_and_locations,eqn:contour_integral_quad_4}.
Computing these weights requires $\approx 10$ matrix-vector multiplies with $\bK$---corresponding to the Lanczos iterations---for a total time complexity of $\bigo{N}$.
All computations involving elliptic integrals can be readily computed using routines available in e.g. the SciPy library.

\begin{algorithm2e}[t!]
  \SetAlgoLined
  \SetKwInOut{Input}{Input}
  \SetKwInOut{Output}{Output}
  \newcommand\NextInput[1]{%
    \settowidth\inputlen{\Input{}}%
    \setlength\hangindent{1.5\inputlen}%
    \hspace*{\inputlen}#1\\
  }
  \newcommand\graycomment[1]{\footnotesize\ttfamily\textcolor{gray}{#1}}
  \SetCommentSty{graycomment}
  \SetKw{Break}{break}
  \SetKwFunction{mvmkxx}{mvm\_$\bK$}
  \SetKwFunction{computequad}{compute\_quad}
  \SetKwFunction{lanczos}{lanczos}
  \SetKwFunction{eig}{symeig}
  \SetKwFunction{ellipke}{ellipke}
  \SetKwFunction{ellipj}{ellipj}
  \SetKwFunction{minres}{msMINRES}
  \SetKwFunction{size}{size}
  \caption{Computing $w_q$ and $t_q$ for Contour Integral Quadrature}
  \label{alg:quadrature}
    \Input{\mvmkxx{$\cdot$} -- function for matrix-vector multiplication (MVM) with matrix $\bK$}
    \NextInput{$Q$ -- number of quad. points}
    \Output{$w_1, \ldots, w_Q$, $t_1, \ldots, t_Q$}
    \BlankLine
    \tcp{Estimate extreme eigenvalues with Lanczos.}
    $\_, \bT$ $\gets$ \lanczos{ \mvmkxx{$\cdot$} } \tcp{Lanczos w/ rand. init. vector}
    $\lambda_\text{min}, \cdots, \lambda_\text{max}$ $\gets$ \eig{$\bT$}
    \BlankLine
    \tcp{Compute elliptic integral of the first kind.}
    \tcp{We use the relation $\mathcal{K}'(k) = \mathcal{K}(k')$, where $k' = \sqrt{1 - k^2}$ is the complementary elliptic modulus.}
    $k^2$ $\gets$ $\lambda_\text{min} / \lambda_\text{max}$ \tcp{The squared elliptic modulus.}
    $k^{\prime 2}$ $\gets$ $\sqrt{1 - k^2}$ \tcp{The squared complementary elliptic modulus.}
    ${\tt K}'$ $\gets$ \ellipke{$k^{\prime 2}$} \tcp{${\tt K}' = \mathcal{K}'(k)$}
    \BlankLine
    \tcp{Compute each quadrature weight/location.}
    \For{$q$ $\gets$ $1$ \KwTo $Q$}{
      $u_q$ $\gets$ $(q - 1/2) / Q$
      \\
      \tcp{Compute Jacobi elliptic fn's via Jacobi's imaginary transform.}
      \tcp{First we compute $\overline{\tt{sn}}_q = \text{sn}(u_q \mathcal{K}'(k) | k')$, $\overline{\tt{cn}}_q = \text{cn}(u_q \mathcal{K}'(k) | k')$, $\overline{\tt{dn}}_q = \text{dn}(u_q \mathcal{K}'(k) | k')$.}
      $\overline{\tt{sn}}_q$, $\overline{\tt{cn}}_q$, $\overline{\tt{dn}}_q$ $\gets$ \ellipj{$u_q {\tt K}'$, $k^{\prime 2}$}
      \\
      \tcp{Use identities to convert $\overline{\tt{sn}}_q$, $\overline{\tt{cn}}_q$, $\overline{\tt{dn}}_q$ values into}
      \tcp{${\tt sn}_q = \text{sn}(i u_q \mathcal{K}'(k) | k)$, ${\tt cn}_q = \text{cn}(i u_q \mathcal{K}'(k) | k)$, ${\tt dn}_q = \text{dn}(i u_q \mathcal{K}'(k) | k)$.}
      ${\tt sn}_q$ $\gets$ $i \left[ \overline{\tt{sn}}_q / \overline{\tt{cn}}_q \right]$
      \\
      ${\tt dn}_q$ $\gets$ $\left[ \overline{\tt{dn}}_q / \overline{\tt{cn}}_q \right]$
      \\
      ${\tt cn}_q$ $\gets$ $\left[ 1 / \overline{\tt{cn}}_q \right]$
      \BlankLine
      \tcp{Quadrature weight $w_q$ and location $t_q$}
      $w_q$ $\gets$ $(-2 \lambda_\text{min}^{1/2})/(\pi Q) \: {\tt K}' \: {\tt cn}_q \: {\tt dn}_q$
      \\
      $t_q$ $\gets$ $\lambda_\text{min} \left( {\tt sn}_q \right)^2$
    }
    \Return{$w_1, \ldots, w_Q$, $t_1, \ldots, t_Q$}
\end{algorithm2e}

\section{The msMINRES Algorithm}
\label{app:minres}

Before introducing the msMINRES algorithm, we will first introduce MINRES as proposed by \citet{paige1975solution}; MINRES can be derived from the Lanczos algorithm \cite{lanczos1950iteration} and, therefore, is able to take advantage of the same three term vector recurrence when building the necessary Krylov subspaces. We will then describe how msMINRES can be derived as a straightforward extension. Notably, we present this section assuming our best initial guess for the linear system we seek to solve is zero. If this is not the case a single step of iterative refinement can be used and the resulting residual system is solved with zero as the initial guess.

\subsection{Standard MINRES}

The method of minimum residuals (MINRES) \cite{paige1975solution} is an alternative to linear conjugate gradients, with the advantage that it can be applied to indefinite and singular symmetric matrices $\bK$.
\citet{paige1975solution} formulate MINRES to solve the least-squares problem $\argmin_{\bc} \Vert \bK \bc - \bb \Vert_2$.
Each iteration $J$ produces a solution $\bc_J$ which is optimal within the $J^\text{th}$ Krylov subspace:
\begin{equation}
	\bc_J^{(\text{MINRES})} = \argmin_{\bc \in \mathcal{K}_J(\bK, \bb)} \Vert \bK \bc - \bb \Vert_2.
	\label{eqn:minres_highlevel}
\end{equation}
Using the Lanczos matrices and some mathematical manipulation, \cref{eqn:minres_highlevel} can be re-formulated as an unconstrained optimization problem:
\begin{align}
  \bc_J^{(\text{MINRES})} &= \Vert \bb \Vert_2 \bQ_J \bz_J
  \nonumber
  \\
  \bz_J &= \argmin_{\by \in \reals^J} \left\Vert
		\left( \widetilde \bT_J \right)\by - \be_1
	\right\Vert_2,
	\quad
  \widetilde \bT_J = \begin{bmatrix} \bT_J \\ \Vert \br_J \Vert_2 \be_J^\top  \end{bmatrix},
  \label{eqn:minres_ols}
\end{align}
where $\be_1, \be_J$ are unit vectors, and $\bQ_J$, $\bT_J$, and $\br_J$ are the outputs from the Lanczos algorithm.
Since \cref{eqn:minres_ols} is a least-squares problem (guaranteed to be full column-rank unless $\bb$ lives in the $J^{\text{th}}$ Krylov subspace---at which point we would exactly solve the problem), we can write the analytic solution to it using the reduced QR factorization of $ \widetilde \bT_J = \bQU_J \bR_J$ \citep[e.g.][]{golub2012matrix}:
\begin{align}
  \bc_J^{(\text{MINRES})} = \Vert \bb \Vert_2 \: \bQ_J \left( \bR^{-1} \bQU_J^\top \right) \be_1.
	\label{eqn:minres_qr}
\end{align}
One way to perform MINRES is first running $J$ iterations of the Lanczos algorithm, computing $\widetilde \bT_J = \bQU_J \bR_J$, and then plugging the resulting $\bQ_J$, $\bQU_J$, and $\bR_J$ into \cref{eqn:minres_qr}.
However, this is unsatisfactory as, na\"ively it requires storing the $N \times J$ matrix $\bQ_J$ \citep[e.g.][]{golub2012matrix} so that $\bc_J$ can be formed.
\citeauthor{paige1975solution} instead introduce a vector recurrence to iteratively compute $\bc_J^\text{(MINRES)}$. This is possible because the QR factorizations of of successive $\widetilde \bT_J$ may be related, allowing for the derivation of a simple update $\bc_{J-1} \rightarrow \bc_{J}$.
This recurrence relation, which is given by \cref{alg:minres} and broadly described below is exactly equivalent to \cref{eqn:minres_qr}; however it uses careful bookkeeping to avoid storing any $N \times J$ terms.

First we note that the $\widetilde \bT_J$ matrices are formed recursively, and thus their QR factorizations are also recursive:
\[
  \bQU^\top \widetilde \bT_J = \begin{bmatrix}
    \bQU_{J-1}^\top & \bqu^{\top(J,1:J-1)} \\
    \bqu^{\top{(1:J-1,J+1)}} & \mathcal{Q}^{(J,J+1)}
  \end{bmatrix} \begin{bmatrix}
    \widetilde \bT_{J-1} & \bt^{(J)} \\ \bzero^\top & \Vert \br_J \Vert
  \end{bmatrix}
  = \begin{bmatrix}
    \bR_{J-1} & {\br^{(J,1:J-1)}} \\
    \bzero & R^{(J,J)}
  \end{bmatrix} = \bR_J
\]
where $\bt^{(J)}$ and $[\br^{(J,1:J-1)}; R^{(J,J)}]$ are the last columns of $\bT_J$ and $\bR_J$ respectively.
Moreover, if we recursively form $\bR_J^{-1}$ as
\[
  \bR_J^{-1} = \begin{bmatrix}
    \bR_{J-1} & {\br^{(J,1:J-1)}} \\
    \bzero & R^{(J,J)}
  \end{bmatrix}^{-1}
  = \begin{bmatrix}
    \bR_{J-1}^{-1} & \left( \bR_{J-1}^{-1} {\br^{(J,1:J-1)}} \right) / R^{(J,J)} \\
    \bzero & 1 / R^{(J,J)}
  \end{bmatrix},
\]
then \cref{eqn:minres_qr} can be re-written in a decent-style update:
\begin{align}
  \bc_J^{(\text{MINRES})} &=
  \Vert \bb \Vert_2 \begin{bmatrix}
    \bQ_{J-1} \bq^{(J)}
  \end{bmatrix} \begin{bmatrix}
    \bR_{J-1}^{-1} & \frac{ \bR_{J-1}^{-1} {\br^{(J,1:J-1)}} }{ R^{(J,J)} } \\
    \bzero & 1 / R^{(J,J)}
  \end{bmatrix} \begin{bmatrix}
    \bQU_{J-1}^\top & \bqu^{\top(J,1:J-1)} \\
    \bqu^{\top{(1:J-1,J+1)}} & \mathcal{Q}^{(J,J+1)}
  \end{bmatrix} \be_1
  \nonumber \\
  &=
  \Vert \bb \Vert_2
  \begin{bmatrix}
    \bQ_{J-1} \bR_{J-1}^{-1} & \frac{ \bQ_{J-1} \bR_{J-1}^{-1} {\br^{(J,1:J-1)}} }{ R^{(J,J)} } \\
    \bzero & 1 / R^{(J,J)} \bq_{J-1}
  \end{bmatrix} \begin{bmatrix}
    \bQU_{J-1}^\top \be_1
    \\
    \mathcal{Q}^{\top{(1,J+1)}}
  \end{bmatrix}
  \nonumber \\
  &=
  \underbracket{\left( \Vert \bb \Vert_2 \bQ_{J-1} \bR_{J-1}^{-1} \bQU_{J-1} \be_1 \right)}_{\bc_{J-1}^{(\text{MINRES})}}
  \:\: + \:\:
  \underbracket{\frac{\Vert \bb \Vert_2 \mathcal{Q}^{\top{(1,J+1)}}}{ R^{(J,J) }}}_{\varphi_J}
  \underbracket{\begin{bmatrix}
    \bQ_{J-1} \bR_{J-1}^{-1} {\br^{(J,1:J-1)}}  \\
    \bq_{J-1}
  \end{bmatrix}}_{\bd_J}.
  \label{eqn:minres_descent}
\end{align}
Thus $\bc^{(\text{MINRES})}_J = \bc^{(\text{MINRES})}_{J-1} + \varphi_J \bd_J$.
The only seemingly expensive part of this update is computing $\bd_J$, as we need to compute $\bQ_{J-1} \bR^{-1}_{J-1} \br^{(J,1:J-1)}$.
$\br^{(J,1:J-1)}$, which is the next entry in the QR factorization of $\widetilde \bT_J$, can be cheaply computed using Givens rotations (see \citep[e.g.][Ch. 11.4.1]{golub2012matrix}).
Moreover, only the last two entries of $\br^{(J,1:J-1)}$ will be non-zero (due to the tridiagonal structure of $\widetilde \bT_J$).
Consequently, we only need to store the last two vectors of $\bQ_{J-1} \bR^{-1}_{J-1}$, which again can be computed recursively.

In total, the whole procedure only requires the storage of $\approx 6$ vectors.
Each iteration requires a single MVM with $\bK$ (to form the next Lanczos vector $\bq_J$); and all subsequent operations are $\bigo{N}$.
The entire procedure is given by \cref{alg:minres}. For simplicity, we have presented the algorithm as if run for a fixed number of steps $J.$ In practice, the MINRES procedure admits inexpensive computation of the residual at each iteration~\cite{paige1975solution} allowing for robust stopping criteria to be used.

\begin{algorithm2e}[t!]
  \SetAlgoLined
  \SetKwInOut{Input}{Input}
  \SetKwInOut{Output}{Output}
  \newcommand\NextInput[1]{%
    \settowidth\inputlen{\Input{}}%
    \setlength\hangindent{1.5\inputlen}%
    \hspace*{\inputlen}#1\\
  }
  \newcommand\graycomment[1]{\footnotesize\ttfamily\textcolor{gray}{#1}}
  \SetCommentSty{graycomment}
  \SetKw{Break}{break}
  \SetKwData{tol}{tolerance}
  \SetKwFunction{mvmkxx}{mvm\_$\bK$}
  \SetKwFunction{mvmprec}{$\bP^{-1}$}
  \SetKwFunction{size}{size}
  \caption[Method of Minimum Residuals (MINRES).]{
    Method of Minimum Residuals (MINRES).
  }
  \label{alg:minres}
    \Input{\mvmkxx{$\cdot$} -- function for MVM with matrix $\bK$}
    \NextInput{$\bb$ -- vector to solve against}
    \Output{$\bc = \bK^{-1} \bb$.}
    \BlankLine
    $\bc_{1}$ $\gets$ $\bzero$ \tcp{Current solution.}
    $\bd_{1}, \bd_{0}$ $\gets$ $\bzero$ \tcp{Current \& prev. ``search'' direction.}
    $\varphi_{2}$ $\gets$ $\Vert \bb \Vert_2$ \tcp{Current ``step'' size.}
    \BlankLine
    \BlankLine
    $\bq_{1}$ $\gets$ $\bb / \Vert \bb \Vert_2$ \tcp{Current Lanczos vector.}
    $\bv_{1}$ $\gets$ \mvmkxx{ $\bq_0$ } \tcp{Buffer for MVM output.}
    $\delta_{1}$ $\gets$ $\Vert \bb \Vert_2$ \tcp{Current Lanczos residual/sub-diagonal.}
    $\delta_{0}$ $\gets$ $1$ \tcp{Prev. Lanczos residual/sub-diagonal.}
    $\eta_{1}$ $\gets$ $1$ \tcp{Current scaling term.}
    $\eta_{0}$ $\gets$ $0$ \tcp{Prev. scaling term.}
    \BlankLine
    \For{$j \gets 2$ \KwTo $J$}{
      \tcp{Run one iter of Lanczos. Gets next vector of $\bQ$ matrix, and next diag/sub-diag ($\gamma$, $\delta$) entries of $\bT$ matrix.}
      $\bq_j$ $\gets$ $\bv_j / \delta_j$
      \\
      $\bv_j$ $\gets$ \mvmkxx{ $\bq_j$ } $ - \delta_j \bq_{j-1} $
      \\
      $\gamma_j$ $\gets$ $\bq_j \bv_j$
      \\
      $\bv_j$ $\gets$ $\bv_j - \gamma_j \bq_j$
      \\
      $\delta_j$ $\gets$ $\Vert \bv_j \Vert$
      \\
      \tcp{Compute the next $\br^{(J)}$ (part of QR) via Givens rotations. There are three non-0 entries: $\bR^{(J,J-2:J)} = [\epsilon_J, \zeta_J, \eta_J]$.}
      $\epsilon_j$ $\gets$ $\delta_{j-1} \left( \delta_{j-2} / \sqrt{\delta_{j-2}^2 + \eta_{j-2}^2} \right)$
      \\
      $\zeta_j$ $\gets$ $\delta_{j-1} \left( \eta_{j-2} / \sqrt{\delta_{j-2}^2 + \eta_{j-2}^2} \right)$
      \\
      $\eta_j$ $\gets$ $\gamma_j \left( \eta_{j-1} / \sqrt{\delta_{j-1}^2 + \eta_{j-1}^2} \right) + \zeta_j \left( \delta_{j-1} / \sqrt{\delta_{j-1}^2 + \eta_{j-1}^2} \right)$
      \\
      $\zeta_j$ $\gets$ $\zeta_j \left( \eta_{j-1} / \sqrt{\delta_{j-1}^2 + \eta_{j-1}^2} \right) + \gamma_j \left( \delta_{j-1} / \sqrt{\delta_{j-1}^2 + \eta_{j-1}^2} \right)$
      \\
      $\eta_j$ $\gets$ $\eta_j \left( \eta_j / \sqrt{\delta_j^2 + \eta_j^2} \right)$
      \\
      \tcp{Compute ``step'' size $\varphi_J = \bQU^{(1,J+1)} / R^{(J,J)} $.}
      $\varphi_{j}$ $\gets$ $\varphi_{j-1} \left( \delta_{j-1} / \sqrt{\delta_{j-1}^2 + \eta_{j-1}^2} \right) \left( \eta_j / \sqrt{\delta_j^2 + \eta_j^2} \right) $
      \\
      \tcp{Update the current solution based on the $\br^{(J)}$ entries ($\epsilon_J, \zeta_J, \eta_J$) and previous search vectors $\bd_{j-1}$, $\bd_{j-2}$.}
      $\bd_j$ $\gets$ $\left( \bq - \zeta_j \bd_{j-1} - \epsilon_j \bd_{j-2} \right) / \eta_j$
      \\
      $\bc_j$ $\gets$ $\bc_{j-1} + \varphi_{j} \bd_j$
    }
    \BlankLine
    \Return{$\Vert \bb \Vert_2 \: \bc_{j}$}
\end{algorithm2e}

\subsection{Multi-Shift MINRES (msMINRES)}

To adapt MINRES to multiple shifts (i.e. msMINRES), we exploit a well-established fact about the shift invariance of Krylov subspaces (see \citep[e.g.][]{datta1991arnoldi,freund1990conjugate,jegerlehner1996krylov,saad2003iterative}).
\begin{observation}
  Let $\bK \bQ_J = \bQ_J \bT_J + \br_J \be_J^\top$ be the Lanczos factorization for $\bK$ given the initial vector $\bb$.
  Then $$(\bK + t \bI) \bQ_J = \bQ_J (\bT_J + t \bI) + \br_J \be_J^\top$$ is the Lanczos factorization for matrix $(\bK + t \bI)$ with initial vector $\bb$.
\end{observation}
\noindent
In other words, if we run Lanczos on $\bK$ and $\bb$, then we get the Lanczos factorization of $(\bK + t \bI)$ \emph{for free}, without any additional MVMs!
Consequently, we can re-use the $\bQ_J$ and $\bT_J$ Lanczos matrices to compute \emph{multiple shifted solves.}
\begin{equation}
  (\bK + t \bI)^{-1} \bb \approx \Vert \bb \Vert_2 \: \bQ_J \left( \bR^{(t){-1}}_J \bQU_J^{(t)\top} \right) \be_1,
  \quad
  \bQU_J^{(t)} \bR_J^{(t)} = \begin{bmatrix} \bT_J + t \bI \\ \Vert \br_J \Vert_2 \be_J^\top \end{bmatrix},
  \label{eqn:minres_qr_shifted}
\end{equation}
Assuming $\bQ$ and $\bT$ have been previously computed, \cref{eqn:minres_qr_shifted} requires no additional MVMs with $\bK$.
We refer to this multi-shift formulation as {\bf Multi-Shift MINRES}, or {\bf msMINRES}.

\newcommand{\blue}[1]{{\color{blue} #1}}
\begin{algorithm2e}[t!]
  \SetAlgoLined
  \SetKwInOut{Input}{Input}
  \SetKwInOut{Output}{Output}
  \newcommand\NextInput[1]{%
    \settowidth\inputlen{\Input{}}%
    \setlength\hangindent{1.5\inputlen}%
    \hspace*{\inputlen}#1\\
  }
  \newcommand\graycomment[1]{\footnotesize\ttfamily\textcolor{gray}{#1}}
  \SetCommentSty{graycomment}
  \SetKw{Break}{break}
  \SetKwData{tol}{tolerance}
  \SetKwFunction{mvmkxx}{mvm\_$\bK$}
  \SetKwFunction{size}{size}
  \caption[Multi-shift MINRES (msMINRES)]{
     Multi-shift MINRES (msMINRES).
     Differences from MINRES (Alg.~\ref{alg:minres}) are in {\color{blue} blue}.
     Blue {\tt for} loops are parallelizable.
  }
  \label{alg:msminres}
    \Input{\mvmkxx{$\cdot$} -- function for MVM with matrix $\bK$}
    \NextInput{$\bb$ -- vector to solve against}
    \NextInput{$t_1, \ldots, t_Q$ -- shifts}
    \Output{$\bc_1 = (\bK + t_1)^{-1} \bb, \ldots, \bc_Q = (\bK + t_Q)^{-1} \bb$.}
    \BlankLine
    $\bq_{1}$ $\gets$ $\bb / \Vert \bb \Vert_2$ \tcp{Current Lanczos vector.}
    $\bv_{1}$ $\gets$ \mvmkxx{ $\bq_0$ } \tcp{Buffer for MVM output.}
    $\delta_{1}$ $\gets$ $\Vert \bb \Vert_2$, $\delta_{0}$ $\gets$ $1$ \tcp{Current/prev. Lanczos residual/sub-diagonal.}
    \For{\color{blue} $q \gets 1$ \KwTo $Q$}{
      $\bc_{1}^{(q)}$ $\gets$ $\bzero$ \tcp{Current solution.}
      $\bd_{1}^{(q)}, \bd_{0}^{(q)}$ $\gets$ $\bzero$ \tcp{Current \& prev. ``search'' direction.}
      $\varphi_{2}^{(q)}$ $\gets$ $\Vert \bb \Vert_2$ \tcp{Current ``step'' size.}
      $\eta_{1}^{(q)}$ $\gets$ $1$, $\eta_{0}^{(q)}$ $\gets$ $0$ \tcp{Current/prev. scaling term.}
    }
    \For{$j \gets 2$ \KwTo $J$}{
      $\bq_j$ $\gets$ $\bv_j / \delta_j$
      \\
      $\bv_j$ $\gets$ \mvmkxx{ $\bq_j$ } $ - \delta_j \bq_{j-1} $
      \\
      $\gamma_j$ $\gets$ $\bq_j \bv_j$
      \\
      $\bv_j$ $\gets$ $\bv_j - \gamma_j \bq_j$
      \\
      $\delta_j$ $\gets$ $\Vert \bv_j \Vert$
      \\
      \For{\color{blue} $q \gets 1$ \KwTo $Q$}{
        $\epsilon_j^{(q)}$ $\gets$ $\delta_{j-1} \left( \delta_{j-2} / \sqrt{\delta_{j-2}^2 + \eta_{j-2}^{(q)2}} \right)$
        \\
        $\zeta_j^{(q)}$ $\gets$ $\delta_{j-1} \left( \eta_{j-2}^{(q)} / \sqrt{\delta_{j-2}^2 + \eta_{j-2}^{(q)2}} \right)$
        \\
        $\eta_j^{(q)}$ $\gets$ $\color{blue} (\gamma_j + t_q) \left( \eta_{j-1}^{(q)} / \sqrt{\delta_{j-1}^2 + \eta_{j-1}^{(q)2}} \right)$ $+ \zeta_j^{(q)} \left( \delta_{j-1} / \sqrt{\delta_{j-1}^2 + \eta_{j-1}^{(q)2}} \right)$
        \\
        $\zeta_j^{(q)}$ $\gets$ $\zeta_j^{(q)} \left( \eta_{j-1}^{(q)} / \sqrt{\delta_{j-1}^2 + \eta_{j-1}^{(q)2}} \right) + $ $\color{blue} (\gamma_j + t_q) \left( \delta_{j-1} / \sqrt{\delta_{j-1}^2 + \eta_{j-1}^{(q)2}} \right)$
        \\
        $\eta_j^{(q)}$ $\gets$ $\eta_j^{(q)} \left( \eta_j^{(q)} / \sqrt{\delta_j^2 + \eta_j^{(q)2}} \right)$
        \\
        $\varphi_{j}^{(q)}$ $\gets$ $\varphi_{j-1}^{(q)} \left( \delta_{j-1} / \sqrt{\delta_{j-1}^2 + \eta_{j-1}^{(q)2}} \right) \left( \eta_j^{(q)} / \sqrt{\delta_j^2 + \eta_j^{(q)2}} \right) $
        \\
        $\bd_j^{(q)}$ $\gets$ $\left( \bq - \zeta_j^{(q)} \bd_{j-1}^{(q)} - \epsilon_j^{(q)} \bd_{j-2}^{(q)} \right) / \eta_j^{(q)}$
        \\
        $\bc_j^{(q)}$ $\gets$ $\bc_{j-1}^{(q)} + \varphi_{j}^{(q)} \bd_j^{(q)}$
      }
    }
    \Return{$\Vert \bb \Vert_2 \: \bc_{j}$}
\end{algorithm2e}

\paragraph{A simple vector recurrence for msMINRES.}
Just as with standard MINRES, \cref{eqn:minres_qr_shifted} can also be computed via a vector recurrence.
We can derive a msMINRES algorithm simply by modifying the existing MINRES recurrence.
Before the QR step in \cref{alg:minres}, we add $t$ to the Lanczos diagonal terms ($\gamma_j + t$, where $\gamma_j = T^{(j,j)}$).
This can be extended to simultaneously handle \emph{multiple shifts} $t_1, \ldots, t_Q$.
Each shift would compute its own QR factorization, its own step size $\varphi_j^{(t_q)}$, and its own search vector $\bd_j^{(t_q)}$.
However, all shifts share the same Lanczos vectors $\bq_j$ and therefore share the same MVMs.
The operations for each shift can be vectorized for efficient parallelization.

To summarize: the resulting algorithm---msMINRES---gives us approximations to $(t_1 \bI + \bK)^{-1} \bb$, $\ldots$, $(t_Q \bI + \bK)^{-1}$ \emph{essentially for free} by leveraging the information we needed anyway to compute $\bK^{-1} \bb$.
\cref{alg:msminres} outlines the procedure; below we re-highlight its computational properties:
\newtheorem*{prop:msminres}{Property~\ref{thm:ciq_convergence} (Restated)}
\begin{prop:msminres}[Computation/Memory of msMINRES-CIQ]
  $J$ iterations of msMINRES requires exactly $J$ matrix-vector multiplications (MVMs) with the input matrix $\bK$,
  regardless of the number of quadrature points $Q$.
  The resulting runtime of msMINRES-CIQ is $\bigo{ J \mvm{\bK}}$, where $\mvm{\bK}$ is the time to perform an MVM with $\bK$.
  The memory requirement is $\bigo{ QN }$ in addition to what's required to store $\bK$.
\end{prop:msminres}

\section{Preconditioning msMINRES-CIQ}
\label{app:preconditioner}

To improve the convergence of \cref{thm:ciq_convergence}, we can introduce a preconditioner $\bP$ where $\bP^{-1} \bK \approx \bI$.
For standard MINRES, applying a preconditioner is straightforward.
We simply use MINRES to solve the system
\[
  \left( \bP^{-1/2} \bK \bP^{-1/2} \right) \bP^{1/2} \bc = \bP^{-1/2} \bb,
\]
which has the same solution $\bc$ as the original system.
In practice the preconditioned MINRES vector recurrence does not need access to $\bP^{-1/2}$---it only needs access to $\bP^{-1}$
(see \citep[][Ch. 3.4]{choi2006iterative} for details).

However, it is not immediately straightforward to apply preconditioning to msMINRES, as preconditioners break the shift-invariance property that is necessary for the $\bigo{JN^2}$ shifted solves \cite{jegerlehner1996krylov,aune2013iterative}.
More specifically, if we apply $\bP$ to msMINRES, then we obtain the solves
\[
	\bP^{-1 / 2} (\bP^{-1 / 2} \bK \bP^{-1 / 2} + t_q \bI)^{-1} (\bP^{-1 / 2} \bb).
\]
Plugging these shifted solves into the quadrature equation \cref{eqn:contour_integral_quad} therefore gives us
\begin{equation}
	\widetilde \ba_J \approx \bP^{-\frac 1 2} (\bP^{-\frac 1 2} \bK \bP^{-\frac 1 2})^{-\frac 1 2} (\bP^{-\frac 1 2} \bb).
  \label{eqn:not_a_sqrt}
\end{equation}
In general, we cannot recover $\bK^{-1/2}$ from \cref{eqn:not_a_sqrt}.
Nevertheless, we can still obtain preconditioned solutions that are equivalent to $\bK^{-1/2} \bb$ and $\bK^{1/2} \bb$ up to an orthogonal rotation.
Let $\bR = \bK \bP^{-1/2} (\bP^{-1/2} \bK \bP^{-1/2})^{-1/2}$.
We have that
\begin{align*}
  \bR \bR^\top
	=
	\bK \left( \bP^{-\frac 1 2} (\bP^{-\frac 1 2} \bK \bP^{-\frac 1 2})^{-\frac 1 2} \right)
	\left( (\bP^{-\frac 1 2} \bK \bP^{-\frac 1 2})^{-\frac 1 2} \bP^{-\frac 1 2} \right) \bK
  = \bK.
\end{align*}
Thus $ \bR $ is equivalent to $\bK^{1/2}$ up to orthogonal rotation.
We can compute $\bR \bb$ (e.g. for sampling) by applying \cref{eqn:not_a_sqrt} to the initial vector $\bP^{1/2} \bb$:
\begin{align}
  \bR \bb
  =
  \bK
  \underbracket{%
    \Bigl[ \bP^{-\frac 1 2} (\bP^{-\frac 1 2} \bK \bP^{-\frac 1 2})^{-\frac 1 2} \bP^{-\frac 1 2} \Bigr]
    \left( \bP^{\frac 1 2} \bb \right)
  }_{\text{Applying preconditioned msMINRES to $\bP^{1/2} \bb$}}.
  \label{eqn:precond_sqrt}
\end{align}
Similarly, $\bR' = \bP^{-1/2} \left( \bP^{-1/2} \bK \bP^{-1/2} \right)^{-1/2}$ is equivalent to $\bK^{-1/2}$ up to orthogonal rotation:
\begin{align*}
  \bR' \bR'^\top
	=
	\left( \bP^{-\frac 1 2} (\bP^{-\frac 1 2} \bK \bP^{-\frac 1 2})^{-\frac 1 2} \right)
	\left( (\bP^{-\frac 1 2} \bK \bP^{-\frac 1 2})^{-\frac 1 2} \bP^{-\frac 1 2} \right)
  = \bK^{-1}.
\end{align*}
We can compute $\bR' \bb$ (e.g. for whitening) via:
\begin{align}
  \bR' \bb
  =
  \underbracket{%
    \Bigl[ \bP^{-\frac 1 2} (\bP^{-\frac 1 2} \bK \bP^{-\frac 1 2})^{-\frac 1 2} \bP^{-\frac 1 2} \Bigr]
    \left( \bP^{\frac 1 2} \bb \right)
  }_{\text{Applying preconditioned msMINRES to $\bP^{1/2} \bb$}}.
  \label{eqn:precond_sqrt_inverse}
\end{align}
Crucially, the convergence of \cref{eqn:precond_sqrt,eqn:precond_sqrt_inverse} depends on the conditioning $\kappa(\bP^{-1} \bK) \ll \kappa(\bK)$.

As with standard MINRES, msMINRES only requires access to $\bP^{-1}$, not $\bP^{-1/2}$.
Note however that \cref{eqn:precond_sqrt,eqn:precond_sqrt_inverse} both require multiplies with $\bP^{1/2}$.
If a preconditioner $\bP$ does not readily decompose into $\bP^{1/2} \bP^{1/2}$, we can simply run the CIQ algorithm on $\bP$ to compute $\bP^{1/2} \bb$.
Thus our requirements for a preconditioner are:
\begin{enumerate}
	\item it affords efficient solves (ideally $o(N^2)$), and
  \item it affords efficient MVMs (also ideally $o(N^2)$) for computing $\bP^{1/2} \bb$ via CIQ.
\end{enumerate}

In our experiments we use the partial pivoted Cholesky preconditioner proposed by \citet{gardner2018gpytorch}, which satisfies the above requirements.
The form of $\bP$ is $\bar \bL \bar \bL^\top + \sigma^2 \bI$, where $\bar \bL$ is a low-rank factor (produced by the partial pivoted Cholesky factorization \cite{harbrecht2012low}) and $\sigma^2 \bI$ is a small diagonal component.
This preconditioner affords $\approx \bigo{N}$ MVMs by exploiting its low rank structure and $\approx \bigo{N}$ solves using the matrix inversion lemma.
Moreover, this preconditioner is highly effective on many Gaussian covariance matrices \cite{gardner2018gpytorch,wang2019exact}.

\section{$\mathcal{O}(M^2)$ Natural Gradient Updates}
\label{app:ngd}

When performing variational inference, we must optimize the $\bmm'$ and $\bS'$ parameters of the whitened variational distribution $q(\bu') = \normaldist{\bmm'}{\bS'}$.
Rather than using standard gradient descent methods on these parameters, many have suggested that {\bf natural gradient descent (NGD)} is better suited for variational inference \cite{hoffman2013stochastic,hensman2012fast,salimbeni2018natural}.
NGD performs the following update:
\begin{equation}
  \begin{bmatrix} \bmm' & \bS' \end{bmatrix} \gets \begin{bmatrix} \bmm' & \bS' \end{bmatrix} - \varphi  \bFS^{-1}
  \begin{bmatrix} \frac{\partial \text{ELBO}}{\partial \bmm'} & \frac{\partial \text{ELBO}}{\partial \bS'} \end{bmatrix}
\end{equation}
where $\varphi$ is a step size, $\begin{bmatrix} \frac{\partial \text{ELBO}}{\partial \bmm'} & \frac{\partial \text{ELBO}}{\partial \bS'} \end{bmatrix}$ is the ELBO gradient, and $\bFS$ is the \emph{Fisher information matrix} of the variational parameters.
Conditioning the gradient with $\bFS^{-1}$ results in descent directions that are better suited towards distributional parameters \cite{hoffman2013stochastic}.

For Gaussian distributions (and other exponential family distributions) the Fisher information matrix does not need to be explicitly computed.
Instead, there is a simple closed-form update that relies on different parameterizations of the Gaussian $\normaldist{\bmm'}{\bS'}$:
\begin{equation}
  \begin{bmatrix} \btheta & \bTheta \end{bmatrix} \gets \begin{bmatrix} \btheta & \bTheta \end{bmatrix} - \varphi
  \begin{bmatrix} \frac{\partial \text{ELBO}}{\partial \boeta} & \frac{\partial \text{ELBO}}{\partial \bEta} \end{bmatrix}.
\end{equation}
$[\btheta, \:\: \bTheta]$ are the Gaussian's \emph{natural parameters}
and $[\boeta, \:\: \bEta]$ are the Gaussian's \emph{expectation parameters}:
\begin{align*}
  \btheta = \bS^{\prime -1} \bmm', &\quad
  \bTheta = -\frac 1 2 \bS^{\prime -1}, \\
  \boeta = \bmm', &\quad
  \bEta = \bmm' \bmm^{\prime\top} + \bS'
\end{align*}

In many NGD implementations, it is common to store the variational parameters via their natural representation ($\btheta$, $\bTheta$), compute the ELBO via the standard parameters ($\bmm'$, $\bS'$), and then compute the derivative via the expectation parameters ($\boeta$, $\bEta$).
Unfortunately, converting between these three parameterizations requires $\bigo{M^3}$ computation.
(To see why this is the case, note that computing $\bS'$ essentially requires inverting the $\bTheta$ matrix.)

\paragraph{A $\bigo{M^2}$ NGD update.}
In what follows, we will demonstrate that the ELBO and its derivative can be computed from $\btheta$ and $\bTheta$ in $\bigo{M^2}$ time via careful bookkeeping.
Consequently, NGD updates have the same asymptotic complexity as the other computations required for SVGP.
Recall that the ELBO is given by
\[
  \text{ELBO} = \overbracket{ \sum_{i=1}^N \Evover{q(f(\bx^{(i)}))}{  \: \log p( y^{(i)} \mid f(\bx^{(i)}) ) \: }}^{\text{expected log likelihood}} - \kl{ q(\bu) }{ p(\bu) }
\]
We will separately analyze the expected log likelihood and KL divergence computations.

\subsection{The Expected Log Likelihood and its Gradient}
Assume we are estimating the ELBO from a single data point $\bx, y$.
The expected log likelihood term of the ELBO is typically computed via Gauss-Hermite quadrature or Monte Carlo integration \cite{hensman2015scalable}:\footnote{
  It can also be computed analytically for Gaussian distributions \cite{hensman2013gaussian}.
  The analytic form achieves the same derivative decomposition as in \cref{eqn:ell_deriv_decomp}
  and so the following analysis will still apply.
}
\[
  \Evover{q(f(\bx)}{ \log p(y \mid f(\bx)) } = \sum_{s=1}^S w_s p(y \mid f_s),
  \quad
  f_s = \ameantest{\bx} + \avartest{\bx}^{1/2} \varepsilon_s
\]
where $w_s$ are the quadrature weights (or $1/S$ for MC integration) and $\varepsilon_s$ are the quadrature locations (or samples from $\normaldist{0}{1}$ for MC integration).
Therefore, the variational parameters only interact with the expected log likelihood term via $\ameantest{\bx}$ and $\avartest{\bx}$.
We can write its gradients via chain rule as:
\begin{align}
  \frac{ \partial \Evover{q(f(\bx)}{ \log p(y | f(\bx)) }}{\partial \boeta}
  \:\: &= \:\:
  c_1 \: \frac{ \partial \ameantest{\bx}}{\partial \boeta}
  \:\: + \:\:
  c_2 \:  \frac{ \partial \avartest{\bx}}{\partial \boeta}
  \nonumber \\
  \frac{ \partial \Evover{q(f(\bx)}{ \log p(y | f(\bx)) }}{\partial \bEta}
  &=
  c_3 \: \frac{ \partial \ameantest{\bx}}{\partial \bEta}
  \:\: + \:\:
  c_4 \:  \frac{ \partial \avartest{\bx}}{\partial \bEta}
  \label{eqn:ell_deriv_decomp}
\end{align}
for some constants $c_1$, $c_2$, $c_3$, and $c_4$ that do not depend on the variational parameters.
It thus suffices to show that the posterior mean/variance and their gradients can be computed from $\btheta$ and $\bTheta$ in $\bigo{M^2}$ time.

\paragraph{The predictive distribution and its gradient.}
All expensive computations involving $\btheta$ and $\bTheta$ are written in {\color{blue} blue}.

$\ameantest{\bx}$ and its derivative can be written as:
\begin{align}
  \ameantest{\bx}
  &= \bk_{\bZ \bx}^\top \bK_{\bZ\bZ}^{-1/2} \bmm'
  \tag{standard parameters} \\
  &= \bk_{\bZ \bx}^\top \bK_{\bZ\bZ}^{-1/2} \boeta
  \tag{expectation parameters} \\
  &= {\color{blue} \bk_{\bZ \bx}^\top \bK_{\bZ\bZ}^{-1/2} (-2 \bTheta)^{-1}} \btheta
  \label{eqn:pred_mean_natural},
  \\
  \frac{\partial \ameantest{\bx}}{\partial \boeta}
  &= \bK_{\bZ\bZ}^{-1/2} \bk_{\bZ \bx}
  \label{eqn:pred_mean_deriv},
  \\
  \frac{\partial \ameantest{\bx}}{\partial \bEta}
  &= \bzero.
  \nonumber
\end{align}

$\avartest{\bx}$ and its derivative can be written as:
\begin{align}
  \avartest{\bx}
  &= \bk_{\bZ \bx}^\top \bK_{\bZ\bZ}^{-1/2} \left( \bS' - \bI \right) \bK_{\bZ\bZ}^{-1/2} \bk_{\bZ \bx}
  \tag{standard parameters} \\
  &= \bk_{\bZ \bx}^\top \bK_{\bZ\bZ}^{-1/2} \left( \bEta - \boeta \boeta^\top - \bI \right) \bK_{\bZ\bZ}^{-1/2} \bk_{\bZ \bx}
  \tag{expectation parameters} \\
  &= {\color{blue} \bk_{\bZ \bx}^\top \bK_{\bZ\bZ}^{-1/2} \left((-2 \bTheta)^{-1} \right.} \left. - \bI \right) \bK_{\bZ\bZ}^{-1/2} \bk_{\bZ \bx}
  \label{eqn:pred_var_natural},
  \\
  \frac{\partial \avartest{\bx}}{\partial \boeta}
  &= -2 \left( {\color{blue} \bk_{\bZ \bx}^\top \bK_{\bZ\bZ}^{-1/2} (-2 \bTheta)^{-1}} \btheta \right) \bK_{\bZ\bZ}^{-1/2} \bk_{\bZ \bx}
  \label{eqn:pred_var_deriv_1},
  \\
  \frac{\partial \avartest{\bx}}{\partial \bEta}
  &= \left( \bK_{\bZ\bZ}^{-1/2} \bk_{\bZ \bx}^\top \right) \left( \bk_{\bZ \bx}^\top \bK_{\bZ\bZ}^{-1/2} \right)
  \label{eqn:pred_var_deriv_2}.
\end{align}
In \cref{eqn:pred_mean_natural,eqn:pred_mean_deriv,eqn:pred_var_natural,eqn:pred_var_deriv_1,eqn:pred_var_deriv_2}, the only expensive operation involving $\bK_{\bZ\bZ}$ is $\bK_{\bZ\bZ}^{-1/2} \bk_{\bZ \bx}$, which can be computed with CIQ.
The only expensive operation involving the variational parameters is ${\color{blue} (-2 \bTheta)^{-1} \bK_{\bZ\bZ}^{-1/2} \bk_{\bZ \bx}}$, which can be computed with preconditioned conjugate gradients after computing $\bK_{\bZ\bZ}^{-1/2} \bk_{\bZ \bx}$.\footnote{
  We typically apply a Jacobi preconditioner to these solves.
}
Those operations only need to be computed once, and then they can be reused across \cref{eqn:pred_mean_natural,eqn:pred_mean_deriv,eqn:pred_var_natural,eqn:pred_var_deriv_1,eqn:pred_var_deriv_2}.
In total, the entire computation for the expected log likelihood and its derivative is $\bigo{M^2}$.

\subsection{The KL Divergence and its Gradient}
We will demonstrate that the KL divergence and its gradient can be computed from $\btheta$ and $\bTheta$ in $\bigo{M^2}$ time.
All expensive computations involving $\btheta$ and $\bTheta$ are written in {\color{blue} blue}.

The whitened KL divergence from \cref{sec:variational_results} is given by:
\begin{align}
  \kl{ q(\bu') }{ p(\bu') }
  &= \frac{1}{2} \left[ \bmm^{\prime \top} \bmm' + \tr{ \bS' } - \log \vert \bS' \vert - M \right]
  \tag{standard parameters} \\
  &= \frac{1}{2} \left[ \tr{ \bEta } - \log \vert \bEta - \boeta \boeta^\top \vert - M \right]
  \tag{expectation parameters} \\
  &= \frac{1}{2} \left[ \btheta^\top {\color{blue} (-2 \bTheta)^{-2} \btheta} + {\color{blue} \tr{(-2 \bTheta)^{-1}}} + {\color{blue} \log \vert -2 \bTheta \vert} - M \right]
  \label{eqn:kl_natural}.
\end{align}
The KL derivative with respect to $\boeta$ and $\bEta$ is surprisingly simple when re-written in terms of the natural parameters
\begin{align}
  \frac{\partial \kl{ q(\bu') }{ p(\bu') }}{ \partial \boeta }
  &= \left( \bEta - \boeta \boeta^\top \right)^{-1} \boeta
  = (\bS')^{-1} \boeta
  \nonumber \\
  &= \btheta
  \label{eqn:kl_mean_deriv}
  \\
  \frac{\partial \kl{ q(\bu') }{ p(\bu') }}{ \partial \bEta }
  &= \frac 1 2 \bI - \frac 1 2 \left( \bEta - \boeta \boeta^\top \right)^{-1}
  = \frac 1 2 \bI - \frac 1 2 (\bS')^{-1}
  \nonumber \\
  &= \frac 1 2 \bI + \bTheta.
  \label{eqn:kl_covar_deriv}
\end{align}
Thus the derivative of the KL divergence only takes $\bigo{M^2}$ time to compute.
The forward pass can also be computed in $\bigo{M^2}$ time---using stochastic trace estimation for the trace term \citep{cutajar2016preconditioning,gardner2018gpytorch}, stochastic Lanczos quadrature for the log determinant \citep{ubaru2017fast,dong2017scalable}, and CG for the solves.
However, during training the forward pass can be omitted as only the gradient is needed for NGD steps.

\section{Experimental Details}
\label{app:experimental_details}

\paragraph{SVGP experiments.}
Each dataset is randomly split into $75\%$ training, $10\%$ validation, and $15\%$ testing sets; $\bx$ and $y$ values are scaled to be zero mean and unit variance.
All models use a constant mean and a Mat\'ern 5/2 kernel, with lengthscales initialized to $0.01$ and inducing points initialized by $K$-means clustering.
Each model is trained for $20$ epochs with a minibatch size of $256.$\footnote{
  The batch size is $512$ on the Covtype dataset due to its larger size.
}
We alternate between optimizing $\bmm'/\bS'$ and the other parameters, using NGD for the former and Adam \cite{kingma2014adam} for the latter.
Each optimizer uses an initial learning rate of $0.01$\footnote{
  On the Precipitation dataset, the initial learning rate is $0.005$ for NGD stability with the Student-T likelihood.
}, decayed by $10\times$ at epochs $1$, $5$, $10$, and $15$.
For CIQ we use $Q = 15$ quadrature points.
msMINRES terminates when the $\bc_j$ vectors achieve a relative norm of $0.001$ or after $J=200$ iterations.
We experimented with tighter tolerances and found no difference in the models' final accuracy.
\new{
  (Note that $J=200$ is almost always enough to achieve the desired $0.001$ tolerance; see \cref{fig:msminres_iters}.)
}
Results are averaged over three trials.

The 3DRoad \citep{guo2012ecomark} and CovType \citep{blackard1999comparative} datasets are available from the UCI repository \citep{asuncion2007uci}.
For 3Droad, we only use the first two features---corresponding to latitude and longitude.
For CovType, we reduce the 7-way classification problem to a binary problem ($\mathtt{Cover\_Type} \in \{ 2, 3\}$ versus $\mathtt{Cover\_Type} \in \{ 0, 1, 4, 5, 6 \}$).
The Precipitation dataset \citep{lyon2004strength,lyon2005enso} is available from the IRI/LDEO Climate Data Library.\footnote{
A processed version of the dataset is available at \url{https://github.com/gpleiss/ciq_experiments/tree/main/svgp/data}.
Original source of data: \url{http://iridl.ldeo.columbia.edu/maproom/Global/Precipitation/WASP_Indices.html}.
}
This spatio-temporal dataset aims to predict the ``WASP'' index (Weighted Anomaly Standardized Precipitation) at various latitudes/longitudes.
Each data point corresponds to the WASP index for a given year (between 2010 and 2019)---which is the average of monthly WASP indices.
In total, there are 10 years and $10,\!127$ latitude/longitude coordinates, for a total dataset size of $101,\!270$.

\paragraph{Bayesian optimization experiments.}
The 6-dimensional Hartmann function is a classical test problem in global optimization\footnote{\url{https://www.sfu.ca/~ssurjano/hart6.html}}.
There are 6 local minima and a global optimal value is $-3.32237$.
We use a total of 100 evaluations with 10 initial points.
The 10 initial points are generated using a Latin hypercube design and we use a batch size of 5.
In each iteration, we draw 5 samples and select 5 new trials to evaluate in parallel.

We consider the same setup and controller as in \cite{eriksson2019scalable} for the 12-dimensional Lunar Lander problem.
The goal is to learn a controller that minimizes fuel consumption and distance to a given landing target while also preventing crashes.
The state of the lunar lander is given by its angle and position, and their time derivatives.
Given this state vector, the controller chooses one of the following four actions: $a \in \{\text{do nothing, booster left, booster right, booster down}\}$.
The objective is the average final reward over a fixed constant set of $50$ randomly generated terrains, initial positions, and initial velocities.
The optimal controller achieves an average reward of $\approx 309$ over the 50 environments.

For both problems, we use a Mat\'ern-$5/2$ kernel with ARD and a constant mean function.
The domain is scaled to \smash{$[0, 1]^d$} and we standardize the function values before fitting the Gaussian process.
The kernel hyperparameters are optimized using L-BFGS-B and we use the following bounds: (lengthscale) $\ell \in [0.01, 2.0\,]$, (signal variance) $s^2 \in [0.05, 50.0]$, (noise variance) $\sigma^2 \in [1e-6, 1e-2]$.
Additionally, we place a horseshoe prior on the noise variance as recommended in \cite{snoek2012practical}.
We add $1\mathrm{e}{-4}$ to the diagonal of the kernel matrix to improve the conditioning and use a preconditioner of rank $200$ for CIQ.

\paragraph{Image reconstruction experiments.}
The matrix $\bA = \bm{D} \bm{B}$ is given as the product of two matrices $\bm{D}$ and $\bm{B}$. Here $\bm{B}$ is a $N^2 \times N^2$ Gaussian
blur matrix with a blur radius of 2.5 pixels and filter size of 5 pixels. The binary matrix $\bm{D}$ is a $KM^2 \times N^2$ downsampling or decimation matrix
that connects the $N \times N$ high-resolution image to the $M \times M$ low-resolution images. For the hyperparameters
$\gamma_{\rm obs}$ and $\gamma_{\rm prior}$ we choose Jeffrey's hyperpriors, i.e.~
\begin{equation}
p(\gamma_{\rm obs}) \propto \gamma_{\rm obs}^{-1} \qquad \qquad {\rm and} \qquad \qquad p(\gamma_{\rm prior}) \propto \gamma_{\rm prior}^{-1}
\end{equation}
In order to conduct the experiment we use the observation likelihood with $\gamma_{\rm obs}=1$ to sample $K=4$ low-resolution images $\by_{1:K}$ from the high-resolution image. The discrete Laplacian matrix
$\bL$ is defined by the following isotropic filter:
\begin{equation}
\bL_{\rm filter}= \frac{1}{12}
  \begin{bmatrix}
    1 & 2    & 1    \\
    2 & -12 & 2    \\
    1 & 2   & 1
  \end{bmatrix}
\end{equation}
For both $\bL$ and $\bm{B}$ we implicitly use reflected (i.e.~non-periodic) boundary conditions.
We use a CG tolerance of $0.001$ and a maximum of $J=400$ msMINRES iterations.
We use a Jacobi preconditioner for CG.
We draw 1000 samples from the Gibbs sampler and treat the first 200 samples as burn-in.
The reconstructed image depicted in the main text is the (approximate) posterior mean. In the main text we provided the conditional posterior for the latent image $\bx$.
To complete the specification of the Gibbs sampler we also need the posterior conditionals for $\gamma_{\rm obs}$ and $\gamma_{\rm prior}$, both of which are given by gamma distributions:
\begin{equation}
\begin{split}
&p(\gamma_{\rm obs} | \bx, \by_{1:K}) = {\rm Ga}(\gamma_{\rm obs} | \alpha =  1 + \tfrac{KM^2}{2}, \beta = 2 / || \by_{1:K} - \bA \bx||^2) \\
&p(\gamma_{\rm prior} | \bx) = {\rm Ga}(\gamma_{\rm prior} | \alpha =  1 + \tfrac{N^2 - 1}{2}, \beta = 2 / || \bL \bx ||^2)
\end{split}
\end{equation}

\section{Proof of Theorem~\ref{thm:ciq_convergence}}
\label{app:proofs}

To prove the convergence result in \cref{thm:ciq_convergence}, we first prove the following lemmas.

\begin{lemma}
  Let $\bK \succ 0$ be symmetric positive definite and let shifts $t_1$, $\ldots$, $t_Q > 0$ be real-valued and positive.
  After $J$ iterations of msMINRES, all shifted solve residuals are bounded by:
  \begin{align*}
    \bigl\Vert (\bK + t_q \bI) \bc_J^{(q)} - \bb \bigr\Vert_2
    &\leq \left( \frac{
      \sqrt{\kappa(\bK + t_q \bI)} - 1
    }{
      \sqrt{\kappa(\bK + t_q \bI)} + 1
    }\right)^J
    \Vert \bb \Vert_2
    \leq \left( \frac{
      \sqrt{\kappa(\bK)} - 1
    }{
      \sqrt{\kappa(\bK)} + 1
    }\right)^J
    \Vert \bb \Vert_2,
	\end{align*}
  where $\bb$ is the vector to solve against, $\bc^{(1)}_J$, $\ldots$, $\bc^{(Q)}$ are the msMINRES outputs, and $\kappa(\bK)$ is the condition number of $\bK$.
  \label{lemma:minres}
\end{lemma}
\begin{proof}
  The convergence proof uses a polynomial bound, which is the standard approach for Krylov algorithms.
  See \citep[e.g.][]{shewchuk1994introduction,trefethen1997numerical,saad2003iterative} for an analogous proof for the conjugate gradients method and~\citep[e.g.][]{greenbaum1997iterative} for a treatment of MINRES applied to both positive definite and indefinite systems.

	At iteration $J$, the msMINRES algorithm produces:
	\begin{align}
    \bc^{(q)}_J
    = \argmin_{\bc^{(q)} \in \mathcal{K}_J(\bK, \bb)} \Bigl[
      \bigl\Vert (\bK + t_q \bI) \bc^{(q)} - \bb \bigr\Vert_2
    \Bigr],
    \quad
    q = 1, \ldots Q,
    \label{eqn:minres_krylov}
	\end{align}
  where without loss of generality we assume $\bc_0^{(q)} = \bzero$ for simplicity.
  Using the fact that Krylov subspaces are shift invariant, we immediately have that
  \begin{align}
    \bc^{(q)}_J
    = \argmin_{\bc^{(q)} \in \mathcal{K}_J(\bK+t_q \bI, \bb)} \Bigl[
      \bigl\Vert (\bK + t_q \bI) \bc^{(q)} - \bb \bigr\Vert_2
    \Bigr],
    \quad
    q = 1, \ldots Q.
  \end{align}
  Since $(\bK + t_q \bI)\succ 0$ we may invoke a result on MINRES error bounds for symmetric positive definite matrices~\cite[Chapter 3]{greenbaum1997iterative} to conclude that
  \[
    \bigl\Vert (\bK + t_q \bI) \bc_J^{(q)} - \bb \bigr\Vert_2
    \leq \left( \frac{
      \sqrt{\kappa(\bK + t_q \bI)} - 1
    }{
      \sqrt{\kappa(\bK + t_q \bI)} + 1
    }\right)^J
    \Vert \bb \Vert_2.
  \]
  Observing that $\kappa(\bK + t_q \bI) \geq \kappa(\bK)$ for all $q$ since $t_q > 0$ concludes the proof.

\end{proof}

\cref{lemma:minres} is a very loose bound, as it doesn't assume anything about the spectrum of $\bK$ (which is standard for generic Krylov method error bounds) and upper bounds the residual error for every shift using the most ill-conditioned system.
In practice, we find that smMINRES converges for many covariance matrices with $J \approx 100$, even when the conditioning is on the order of $\kappa(\bK) \approx 10^4$ and this convergence can be further improved with preconditioning.

\begin{lemma}
  For any positive definite $\bK$ and positive $t$, we have
  \begin{align}
    \frac{
      \sqrt{\kappa(\bK + t \bI)} - 1
    }{
      \sqrt{\kappa(\bK + t \bI)} + 1
    } = \frac{\sqrt{\lambda_\text{max} + t} - \sqrt{\lambda_\text{min} + t}  }{\sqrt{\lambda_\text{max} + t} + \sqrt{\lambda_\text{min} + t}  }
    < \frac{\lambda_\text{max}}{4t}
  \end{align}
  \label{lemma:condition}
\end{lemma}

\begin{proof}
  We can upper bound the numerator
  \begin{align*}
    \sqrt{\lambda_\text{max} + t} - \sqrt{\lambda_\text{min} + t}
    &\leq
    \sqrt{\lambda_\text{max} + t} - \sqrt{t}
    \\
    &=
    \sqrt{\lambda_\text{max}} \left( \sqrt{1 + t/\lambda_\text{max}} - \sqrt{t/\lambda_\text{max}} \right)
    \leq
    \sqrt{\lambda_\text{max}} \frac{1}{2 \sqrt{t/\lambda_\text{max}}}
    =
    \frac{\lambda_\text{max}}{2 \sqrt{t}}.
  \end{align*}
  where we have applied the standard inequality $\sqrt{(\cdot)+1} - \sqrt{(\cdot)} < \frac{1}{2 \sqrt{(\cdot)}}$.
  The denominator can be (loosely) lower-bounded as $2\sqrt{t}$.
  Combining these two bounds completes the proof.
\end{proof}

\begin{lemma}
  Let $\sigma_q^2$ and $\widetilde w_q$ be defined as in \cref{eqn:quad_points_and_locations}.
  Then
  \begin{equation*}
    \sum_{q=1}^Q \frac{|w_q|}{|t_q|} = \sum_{q=1}^Q \frac{|\widetilde w_q|}{|\sigma^2_q|} < \frac{4 Q \log \left( 5 \sqrt{\kappa(\bK)} \right)  }{\pi \sqrt{\lambda_\text{min}}} \\
  \end{equation*}
  where $w_q = -\widetilde w_q$ and $t_q = -\sigma^2_q$ as used in \cref{eqn:contour_integral_quad_4}.
  \label{lemma:quad_ratio}
\end{lemma}

\begin{proof}
  Using facts about elliptical integrals we have
  \begin{align}
    \mathcal{K}'(k) < \log(1 + 4 / k) \leq \log( 5 / k) &\qquad k \in (0, 1)
    \tag{\citep[][Thm. 1.7]{qiu1998some} and \citep[][Thm. 2]{yang2019convexity}}
    \\
    \frac{\pi}{2} \leq \mathcal{K}(k) &\qquad k\in[0,1] \tag{\citep[e.g.][]{qiu1998some}}
  \end{align}
  where in the first statement we have used that $\mathcal{K}'(k) = \mathcal{K}(k').$ For Jacobi elliptic functions we have that
  \begin{align}
    0 < \text{dn}(u \mathcal{K}(k)| k ) < 1 &\qquad u\in (0,1), \; k \in (0, 1) \tag{\citep[e.g.][]{meyer2001jacobi}}
    \\
    0 < \text{sn}( u \mathcal{K}(k)| k ) < 1 &\qquad u\in(0,1), \; k \in (0, 1) \tag{\citep[e.g.][]{meyer2001jacobi}}
    \\
    \text{sn}( \pi u /2 | 0 ) < \text{sn}( u \mathcal{K}(k)| k ) < 1 &\qquad u\in(0,1), \; k \in (0, 1) \tag{\citep[][Thm. 1]{carlson1983degenerating}}
  \end{align}
  where in the last inequality we have used that $\mathcal{K}(0) = \pi / 2$~\citep[e.g.][]{abramowitz1948handbook}. Coupling the final inequality above with $\text{sn}( \pi u /2 | 0 ) = \sin(\pi u / 2)$ for $u\in(0,1)$ we have that
  \[
  \sin( \pi u /2 ) < \text{sn}( u \mathcal{K}(k)| k ) < 1 \qquad u\in(0,1), \; k \in (0, 1).
  \]

  Now, for each $q$ we have that
  \begin{align*}
  \frac{w_q}{t_q}
    =
    \frac{\widetilde w_q}{\sigma^2_q}
    &= \left(\frac{- 2\sqrt{\lambda_{\min}}}{\pi Q \lambda_{\min}}\right)\frac{\mathcal{K}'( k )\text{cn} \left( i u_q \mathcal{K}'(k) \mid k \right)\text{dn} \left( i u_q \mathcal{K}'(k) \mid k \right)}{\text{sn}(i u_q \mathcal{K}'(k) \mid k)^2}
    \\
    &=\left(\frac{2\mathcal{K}'( k )}{\pi Q \lambda_{\min}}\right)\frac{\text{dn} \left( u_q \mathcal{K}(k') \mid k' \right)}{\text{sn}(u_q \mathcal{K}(k') \mid k')^2}
    \tag{via Jacobi imaginary transforms~\citep[e.g.][]{abramowitz1948handbook}}
  \end{align*}

  Consequently, we may conclude that
  \begin{align*}
    \frac{|w_q|}{|t_q|}
    &= \left(\frac{2\mathcal{K}'( k )}{\pi Q \lambda_{\min}}\right)\frac{\text{dn} \left( u_q \mathcal{K}(k') \mid k' \right)}{\text{sn}(u_q \mathcal{K}(k') \mid k')^2} \\
    &\leq \frac{2\log( 5 / k)}{\pi Q \lambda_{\min}}\left(\frac{1}{\sin^2( \pi u_q /2 )} \right)
  \end{align*}
  where we note that all quantities on the right hand side are positive.
  Plugging in the values of $k = 1/\sqrt{\kappa{(\bK)}}$, $u_q = (q-1/2)/Q$ and summing over $u_q$ we see that
  \begin{align}
    \sum_{q=1}^Q \frac{|w_q|}{|t_q|}
    &<
    \sum_{q=1}^Q \frac{ 2 \log \left( 5 \sqrt{\kappa(\bK)} \right)}
    { \pi Q \sqrt{\lambda_\text{min}} \sin^2 ( \frac{\pi (q - 1/2)}{2Q}) }.
  \end{align}
  Through trigonometric identities
  $\sum_{q=1}^Q 1 / ( Q \sin^2 \frac{\pi(q- 1/2 )}{2Q} )  = 2 Q$ and, therefore,
  \begin{align*}
    \sum_{q=1}^Q \frac{|w_q|}{|t_q|}
    &< \frac{4 Q \log \left( 5 \sqrt{\kappa(\bK)} \right)}{\pi \sqrt{\lambda_\text{min}}}.
  \end{align*}
\end{proof}

With these lemmas we are now able to prove Theorem~\ref{thm:ciq_convergence}:
\newtheorem*{ciq_convergence}{Theorem~\ref{thm:ciq_convergence} (Restated)}
\begin{ciq_convergence}
  Let $\bK \succ 0$ and $\bb$ be inputs to msMINRES-CIQ, producing $\ba_J \approx \bK^{1/2} \bb$ after $J$ iterations with $Q$ quadrature points.
  The difference between $\ba_J$ and $\bK^{1/2} \bb$ is bounded by:
  \begin{equation*}
    \left\Vert \bv_J - \bK^{\frac 1 2} \bb \right\Vert_2
    \leq
    \overbracket{
      \bigo{\exp\left( -\tfrac  {2 Q \pi^2}{\log \kappa(\bK) + 3} \right)}
    }^{\text{Quadrature error}}
    +
    \overbracket{
      \tfrac{ 2 Q \log \left( 5 \sqrt{\kappa(\bK)} \right) \kappa(\bK) \sqrt{\lambda_\text{min}} }{\pi}
      \left( \tfrac{ \sqrt{\kappa(\bK)} - 1}{ \sqrt{\kappa(\bK)} + 1} \right)^{J-1}
      \left\Vert \bb \right\Vert_2.
    }^{\text{msMINRES error}}
  \end{equation*}
  where $\lambda_\text{max},\lambda_{\text{min}}$ are the max and min eigenvalues of $\bK$, and $\kappa(\bK)$ is the condition number of $\bK$.
\end{ciq_convergence}
\begin{proof}
  First we note that the msMINRES-CIQ solution $\ba_J$ can be written as $\sum_{i=1} w_q \bc^{(q)}_J$, where $\bc^{(q)}_J$ is the $q^\text{th}$ shifted solve $\approx (t_q \bI + \bK)^{-1} \bb$ from msMINRES.
  Applying the triangle inequality we have:
  \begin{align}
    \left\Vert \ba_J - \bK^{\frac 1 2} \bb \right\Vert_2
    &=
    \left\Vert \overbracket{\sum_{q=1}^Q w_q \bc^{(q)}_J - \left( \bK \sum_{q=1}^Q w_q \left( t_q \bI + \bK \right)^{-1} \right) \bb }^{\text{msMINRES error}} \right.
    \nonumber
    \\
    &\phantom{=} \quad \left. + \underbracket{\left( \bK \sum_{q=1}^Q w_q \left( t_q \bI + \bK \right)^{-1} \right) \bb - \bK^{\frac 1 2} \bb}_{\text{Quadrature error}} \right\Vert_2
    \nonumber
    \\
    &\leq \sum_{q=1}^Q \vert w_q \vert \left\Vert \bc^{(q)}_J - \bK \left( t_q \bI + \bK \right)^{-1} \bb \right\Vert_2
    \nonumber
    \\
    &\phantom{=} \:\: + \left\Vert \bK \left( \sum_{q=1}^Q w_q \left( t_q \bI + \bK \right)^{-1} \right) \bb - \bK^{\frac 1 2} \bb \right\Vert_2
    \label{eqn:pre_bound}
  \end{align}
  Plugging \cref{lemma:minres} into the msMINRES part of the bound bound, we have:
  \begin{align*}
    \sum_{q=1}^Q \left\vert w_q \right\vert&
    \left( \frac{ \sqrt{\kappa(\bK + t_q \bI)} - 1}{ \sqrt{\kappa(\bK + t_q \bI)} + 1} \right)^J \left\Vert \bb \right\Vert_2
    \\
    \leq \:\:
    & \sum_{q=1}^Q \left\vert w_q \right\vert
    \left( \frac{ \sqrt{\kappa(\bK + t_q \bI)} - 1}{ \sqrt{\kappa(\bK + t_q \bI)} + 1} \right)
    \left( \frac{ \sqrt{\kappa(\bK)} - 1}{ \sqrt{\kappa(\bK)} + 1} \right)^{J-1}
    \left\Vert \bb \right\Vert_2
    \tag{via \cref{lemma:minres}}
    \\
    \leq \:\:
    & \sum_{q=1}^Q \left\vert w_q \right\vert
    \left( \frac{ \lambda_\text{max} }{ 4 t_q } \right)
    \left( \frac{ \sqrt{\kappa(\bK)} - 1}{ \sqrt{\kappa(\bK)} + 1} \right)^{J-1}
    \left\Vert \bb \right\Vert_2
    \tag{via \cref{lemma:condition}}
    \\
    \leq \:\:
    & \frac{ 2 Q \log \left( 5 \sqrt{\kappa(\bK)} \right) \lambda_\text{max}}{\pi \sqrt{\lambda_\text{min}}}
    \left( \frac{ \sqrt{\kappa(\bK)} - 1}{ \sqrt{\kappa(\bK)} + 1} \right)^{J-1}
    \left\Vert \bb \right\Vert_2
    \tag{via \cref{lemma:quad_ratio}}
    \\
    \leq \:\:
    & \frac{ 2 Q \log \left( 5 \sqrt{\kappa(\bK)} \right) \sqrt{\lambda_{\min}}\kappa(\bK)}{\pi}
    \left( \frac{ \sqrt{\kappa(\bK)} - 1}{ \sqrt{\kappa(\bK)} + 1} \right)^{J-1}
    \left\Vert \bb \right\Vert_2.
  \end{align*}
   Plugging this bound and \cref{lemma:hale} into \cref{eqn:pre_bound} completes the proof.
\end{proof}

We can also prove this simple corollary:
\begin{corollary}
  Let $\bK\succ 0$ and $\bb$ be the inputs to \cref{alg:ciq}, producing the output $\ba_J' \approx \bK^{-1/2} \bb$ after $J$ iterations with $Q$ quadrature points.
  The difference between $\ba_J$ and $\bK^{1/2} \bb$ is bounded by:
  \begin{equation*}
    \left\Vert \ba_J' - \bK^{-\frac 1 2} \bb \right\Vert_2
    \leq
    \overbracket{
      \bigo{ \tfrac{1}{\lambda_\text{min}} \exp\left( -\tfrac  {2 Q \pi^2}{\log \kappa(\bK) + 3} \right)}
    }^{\text{Quadrature error}}
    +
    \overbracket{
      \tfrac{ 2 Q \log \left( 5 \sqrt{\kappa(\bK)} \right) \kappa(\bK) }{ \sqrt{\lambda_\text{min}} \pi}
      \left( \tfrac{ \sqrt{\kappa(\bK)} - 1}{ \sqrt{\kappa(\bK)} + 1} \right)^{J-1}
      \left\Vert \bb \right\Vert_2.
    }^{\text{msMINRES error}}
  \end{equation*}
  where $\lambda_\text{max},\lambda_{\text{min}}$ are the maximal and minimal eigenvalues of $\bK$, and $\kappa(\bK)$ is the condition number of $\bK$.
  \label{thm:ciq_convergence_inverse}
\end{corollary}

\begin{proof}
  Note that $\ba_J' = \bK^{-1} \ba_J$, where $\ba_J$ is the msMINRES-CIQ estimate of $\bK^{1/2} \bb$.
  Using the sub-multiplicative property of the induced matrix 2-norm we see that
  \[
    \left\Vert \ba_J' - \bK^{-\frac 1 2} \bb \right\Vert_2
    \leq \left\Vert \bK^{-1} \right\Vert_2 \left\Vert \ba_J - \bK^{\frac 1 2} \bb \right\Vert_2
    = \frac{1}{\lambda_\text{min}} \left\Vert \ba_J - \bK^{\frac 1 2} \bb \right\Vert_2,
  \]
  where the final term is bounded by \cref{thm:ciq_convergence}.
\end{proof}

\end{document}